\documentclass[10pt,twocolumn,letterpaper]{article}

\usepackage{cvpr}
\usepackage{times}
\usepackage{epsfig}
\usepackage{graphicx}
\usepackage{amsmath}
\usepackage{amssymb}

\usepackage{algorithm}
\usepackage{algorithmic}
\usepackage{amsfonts}
\usepackage{amsthm}  
\usepackage{wasysym}
\usepackage[ruled,vlined,algo2e]{algorithm2e}
\providecommand{\SetAlgoLined}{\SetLine}
\usepackage{color}
\usepackage{dsfont}	
\usepackage{wrapfig}
\usepackage{footnote}
\usepackage{epstopdf}

\newtheorem*{thm*}{Theorem}

\newtheorem*{definition*}{Definition}
\newtheorem*{cor*}{Corollary}

\newcommand{\rmnum}[1]{\romannumeral #1}

\DeclareMathOperator*{\argmin}{arg\,min}
\DeclareMathOperator*{\argmax}{arg\,max}
\def\decide#1#2{
  \mathrel{
    \mathop{
      \begin{array}{c}
        >\vspace{-1.4ex}\\<
      \end{array}
      }\limits_{#2}\limits^{#1}
    }
  }

\newcommand{\resp}{{\it resp. }}

\usepackage[pagebackref=true,breaklinks=true,letterpaper=true,colorlinks,bookmarks=false]{hyperref}

\cvprfinalcopy 

\def\cvprPaperID{1675} 

\ifcvprfinal\fi
\begin{document}

\title{Zero-Shot Learning via Joint Latent Similarity Embedding}

\author{Ziming Zhang and Venkatesh Saligrama\\
Department of Electrical \& Computer Engineering, Boston University\\
{\tt\small \{zzhang14, srv\}@bu.edu}
}

\maketitle

\begin{abstract}
Zero-shot recognition (ZSR) deals with the problem of predicting class labels for target domain instances based on source domain side information (\eg attributes) of unseen classes. We formulate ZSR as a binary prediction problem. Our resulting classifier is class-independent. It takes an arbitrary pair of source and target domain instances as input and predicts whether or not they come from the same class, \ie whether there is a match. We model the posterior probability of a match since it is a sufficient statistic and propose a latent probabilistic model in this context. We develop a joint discriminative learning framework based on dictionary learning to jointly learn the parameters of our model for both domains, which ultimately leads to our class-independent classifier. 
Many of the existing embedding methods can be viewed as special cases of our probabilistic model. On ZSR our method shows 4.90\% improvement over the state-of-the-art in accuracy averaged across four benchmark datasets. We also adapt ZSR method for zero-shot retrieval and show 22.45\% improvement accordingly in mean average precision (mAP). 
\end{abstract}
\section{Introduction}
Zero-shot learning (ZSL) deals with the problem of learning to classify previously unseen class instances. It is particularly useful in large scale classification where labels for many instances or entire categories can often be missing. One popular version of ZSL is based on the so-called source and target domains. In this paper we consider the source domain as a collection of class-level vectors, where each vector describes side information of one {\em single} class with, for instance, attributes \cite{farhadi2009attribute,10.1109/TPAMI.2013.140,mensink2012metric,Parikh:2011:IBD:2191740.2191861,rohrbach2011largeScale}, language words/phrases \cite{Berg:2010:AAD:1886063.1886114,frome2013devise,socher2013zero}, or even learned classifiers \cite{yu2013designing}. The target domain is described by a distribution of instances (\eg images, videos, {\it etc.}) \cite{10.1109/TPAMI.2013.140, wu2014zero}. During training, we are given source domain side information and target domain data corresponding to only a subset of classes, which we call {\em seen} classes. During test time for the source domain, side information is then provided for {\em unseen} classes. A target domain instance from an unknown {\em unseen} class is then presented. The goal during test time is to predict the class label for the unseen target domain instance. 
%

\begin{figure}[t]
\centerline{\includegraphics[width=\linewidth]{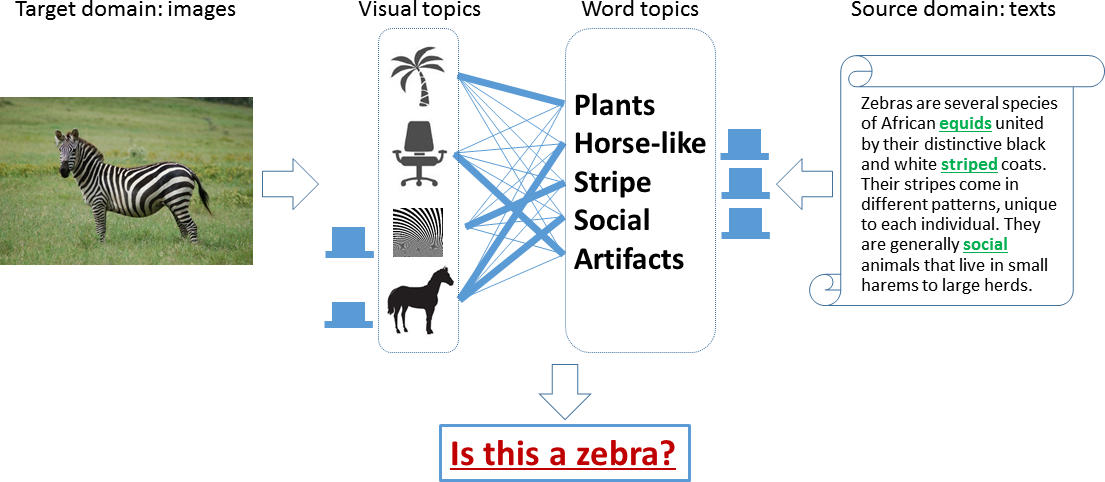}}
\vspace{1mm}
\caption{\footnotesize{Illustration of our joint latent space model with images as target domain and text-documents as source domain. The bar graph next to the (latent) topics indicate the mixture weights of the topics. The links between the topics indicate the co-occurrence (thicker lines depicting larger likelihood values). Our method is based on learning a class-independent similarity function using seen class training data, which measures the likelihood of a source domain class vector and a target domain data sample being the same class, regardless of their true underlying classes.}}\label{fig:rep}
\vspace{-3mm}
\end{figure}

\noindent
{\bf Intuition:} In contrast to previous methods (\eg \cite{Akata2015}) which explicitly learn the relationships between source and target domain data, we posit that for both domains there exist corresponding latent spaces, as illustrated in Fig. \ref{fig:rep}, where there is a similarity function {\em independent} of class labels. 

Our supposition implies that, regardless of the underlying class labels, there is a statistical relationship between latent co-occurrence patterns of corresponding source and target instance pairs when the instance pairs describe the same thing. For example, with our supposition the ``zebra'' image in Fig.~\ref{fig:rep} on the left will share an underlying statistical relationship with the description of zebra in text on the right, and that this relationship can be inferred by means of a class-independent ``universal'' similarity function\footnote{Intuitively this is a plausible mechanism. We as humans tend to draw connections from different sources to improve our understanding of objects/concepts.}.

To mathematically formalize this intuition we formulate zero-shot recognition (ZSR) as a binary classification problem. In this framework, we train a score function that takes an arbitrary source-target instance pair as input and outputs a likelihood score that the paired source and target instances come from the same class. We apply this score function on a given target instance to identify a corresponding source vector with the largest score. In this way our score function generalizes to unseen classes since it does not explicitly depend on the actual class label. 

We train our binary predictor (\ie score function) using seen class source and target domain data. It is well-known that for a binary classification problem the posterior probability of the binary output conditioned on data is a sufficient statistic for optimal detection. This motivates us to propose a latent parametrized probabilistic model for the posterior. We decompose the posterior into source/target domain data likelihood terms and a cross-domain latent similarity function. We develop a joint discriminative learning framework based on dictionary learning to jointly learn the parameters of the likelihood and latent similarity functions. 

In test-time unseen source domain vectors are revealed. We estimate their corresponding latent source embeddings. Then, for an arbitrary target-instance, we estimate the latent target embedding. Finally we score each pair of source and target domain embeddings using our similarity function and classify based on these scores. Fig. \ref{fig:rep} illustrates a specific scenario where visual and word embedding functions are learned using training data from seen classes and are utilized to estimate embeddings for unseen data. 
We test our method on four challenging benchmark datasets (\ie aP\&Y, AwA, CUB, SUN-attribute). Our performance on average shows 4.9\% improvement in recognition accuracy. We also adapt ZSR method for zero-shot retrieval and show 22.45\% improvement in mean average precision across these datasets. 

Our proposed general probabilistic model is a systematic framework for ZSR. Indeed, existing methods including \cite{akata2013label,Akata2015,frome2013devise,hariharan2012efficient,mensink2014costa,norouziMBSSFCD14} can be precisely interpreted as special cases of our method.  
We test our algorithm on several ZSL benchmark datasets and achieve state-of-the-art results. 

\subsection{Related Work}
\noindent
{\bf (\rmnum{1}) Attribute prediction:} A significant fraction of zero-shot methods are based on building attribute classifiers that transfer target domain data into source domain attribute space. For instance, \cite{palatucci2009zero} used semantic knowledge bases to learn the attribute classifiers. \cite{10.1109/TPAMI.2013.140, mahajan2011joint, wang2013unified, yu2013designing, yu2010attribute} proposed several (probabilistic or discriminative) attribute prediction methods using the information from attributes, classes, and objects. \cite{mensink2014costa} proposed combining seen class classifiers linearly to build unseen class classifiers. \cite{hariharan2012efficient} proposed first linearly projecting both source and target domain data into a common space and then training a max-margin multi-label classifiers for prediction. \cite{Romera-Paredes2015} proposed a related regularization based method for training classifiers. The main issue in such methods is that they may suffer from noisy source/target data, which often results in poor prediction. In contrast, our joint latent space model is robust to the noise issues on account of the nature of latent space learning.

\noindent
{\bf (\rmnum{2}) Linear embedding:} This type of methods are based on embedding both source and target domain data into a feature space characterized by the Kronecker product of source domain attributes and target domain features. Linear classifiers are trained in the product space. For instance, \cite{akata2013label} created such spaces using label embedding, and \cite{Akata2015,frome2013devise,norouziMBSSFCD14,socher2013zero} utilized deep learning for the same purpose. Recently \cite{Li2015, Li_ICCV2015} introduced semi-supervised max-margin learning to learn the label embedding.

\noindent
{\bf (\rmnum{3}) Nonlinear embedding:} Similar to linear embedding, here the Kronecker product feature space is constructed after a nonlinear mapping of the original features. This literature includes \cite{Ba2015, Kodirov2015, Zhang2015}, where \cite{Kodirov2015, Zhang2015} embed source and target domain data nonlinearly into known semantic spaces (\ie seen classes) in an unsupervised or supervised way, and \cite{Ba2015} employed deep neural networks for associating the resulting embeddings. 

Different from these (linear or nonlinear) embedding based zero-shot methods, our method learns a {\em joint latent space} for both domains using {\em structured learning}. The learned joint space is used not only to fit each instance well (by dictionary learning) but also to enable recognition (by bilinear classifiers) during test time. 

\noindent
{\bf (\rmnum{4}) Other methods:} Less related to our method includes approaches based on semantic transfer propagation \cite{conf/nips/RohrbachES13}, transductive multi-view embedding \cite{embedding2014ECCV}, random forest approach \cite{jayaraman2014unreliable}, and semantic manifold distance \cite{fu2015}. 

\section{Our Method}
\subsection{Problem Setting}
Let us motivate our approach from a probabilistic modelling perspective. This will in turn provide a basis for structuring our discriminative learning method. We denote by ${\cal X}^{(s)}$ the space of source domain vectors, by ${\cal X}^{(t)}$ the space of target domain vectors, and  by ${\cal Y}$ the collection of all classes. 
Following convention, the random variables are denoted by capital letters, namely, $X^{(s)},\,X^{(t)},Y$ and instances of them by lower-case letters $\mathbf{x}^{(s)}, \mathbf{x}^{(t)}, \mathbf{y}$. 

Zero-shot learning is a special case where the class corresponding to the source domain instance is revealed during test time and thus there is no uncertainty regarding the class label for any source domain vector. Thus the problem reduces to assigning target domain instances to source domain vectors (and in turn to classes) during testing. For exposition we denote by $y^{(s)}$ the label for the source domain instance $\mathbf{x}^{(s)} \in {\cal X}^{(s)}$ even though we know that $y^{(s)}$ is identical to the true class label $y$. With this in mind, we predict a class label $y^{(t)}$ for target domain instance $\mathbf{x}^{(t)} \in {\cal X}^{(t)}$. 
 
%
%

\subsection{General Probabilistic Modelling}
Abstractly, we can view ZSR as a problem of assigning a {\em binary} label to a pair of source and target domain instances, namely whether or not $y^{(st)}\triangleq [y^{(s)} = y^{(t)}]$ holds. 

We view our goal in terms of evaluating how likely this proposal is true, \ie $p(y^{(st)} | \mathbf{x}^{(s)}, \mathbf{x}^{(t)})$. Indeed, Bayes Optimal Risk theory tells us that the optimal classifier (see Eq.~6 in \cite{duda1995pattern}), $F(\mathbf{x}^{(s)}, \mathbf{x}^{(t)})$, is obtained by suitably thresholding the posterior of $y^{(st)}$ conditioned on data, namely,
\begin{align}
F(\mathbf{x}^{(s)}, \mathbf{x}^{(t)}) \triangleq \log p(y^{(st)} | \mathbf{x}^{(s)}, \mathbf{x}^{(t)}) \decide{\mbox{Ident}}{\mbox{Diff}} \theta
\end{align}
where $\theta \in \mathbb{R}$ is a threshold parameter. Here {\it Ident} is the hypothesis that source/target data describe the same class. {\it Diff} is the hypothesis that they are different. 

Our latent embedding model supposes that the observed and latent random variables form a Markov chain \cite{cover}: 
\begin{align} \label{e.markov}
X^{(s)} \leftrightarrow Z^{(s)} \leftrightarrow Y  \leftrightarrow Z^{(t)} \leftrightarrow X^{(t)}.
\end{align}
This implies that the source domain data, $X^{(s)}$, and its associated embedding, $Z^{(s)}$ is independent of the target $X^{(t)}, Z^{(t)}$ conditioned on the underlying class $Y$ (if they belong to the same class) and unconditionally independent if they belong to different classes.  

It follows that the posterior probability can be factored as
$p(y^{(st)}, \mathbf{z}^{(s)}, \mathbf{z}^{(t)} | \mathbf{x}^{(s)}, \mathbf{x}^{(t)}) = p(y^{(st)} | \mathbf{z}^{(s)}, \mathbf{z}^{(t)}) p(\mathbf{z}^{(s)}, \mathbf{z}^{(t)} | \mathbf{x}^{(s)}, \mathbf{x}^{(t)})$.
Next note that, in the absence of class information, it is reasonable to assume that an arbitrary pair of source and target domain latent embeddings are essentially independent, namely,  $p(\mathbf{z}^{(s)}, \mathbf{z}^{(t)}) \approx p(\mathbf{z}^{(s)})p(\mathbf{z}^{(t)})$. 
Consequently, the posterior probability can be expressed as follows:
\begin{align} \label{eqn:p} 
& p( y^{(st)} | \mathbf{x}^{(s)}, \mathbf{x}^{(t)}) \\ 
& = \sum_{\mathbf{z}^{(s)}, \mathbf{z}^{(t)}} p(\mathbf{z}^{(s)}|\mathbf{x}^{(s)})p(\mathbf{z}^{(t)}|\mathbf{x}^{(t)})p(y^{(st)}|\mathbf{z}^{(s)}, \mathbf{z}^{(t)}), \nonumber
\end{align}
where, $\mathbf{z}^{(s)} \in \mathbb{R}^{h_s}$ and $\mathbf{z}^{(t)} \in \mathbb{R}^{h_t}$ denote the latent coefficient vectors in the corresponding $h_s$-dim and $h_t$-dim latent spaces, respectively. Here $(\mathbf{z}^{(s)},\mathbf{z}^{(t)})$ defines the {\em joint latent embedding} for data pair $(\mathbf{x}^{(s)},\mathbf{x}^{(t)})$. This factorization provides us two important insights:

\underline{\em (\rmnum{1}) Class-independent Embeddings:} Note that the expression in Eq.~\ref{eqn:p} informs us that the probability kernels 
$p(\mathbf{z}^{(s)}|\mathbf{x}^{(s)}),\,\,p(\mathbf{z}^{(t)}|\mathbf{x}^{(t)})$ characterizing the latent embeddings depend only on the corresponding  data instances, $\mathbf{x}^{(s)},\,\mathbf{x}^{(t)}$ and independent of the underlying class labels.

\underline{\em (\rmnum{2}) Class-independent Similarity Kernel:} The expression in Eq.~\ref{eqn:p} reveals that the term $p(y^{(st)}|\mathbf{z}^{(s)}, \mathbf{z}^{(t)})$ is a class-invariant function that takes arbitrary source and target domain embeddings as input and outputs a likelihood of similarity regardless of underlying class labels (recall that predicting $y^{(st)}\triangleq[y^{(s)}=y^{(t)}]$ is binary).
%
%
Consequently, at a conceptual level, our framework  provides a way to assign similarities of class membership between arbitrary target domain vectors and source domain vectors while circumventing the intermediate step of assigning class labels.

In our context the joint probability distributions and latent conditionals are unknown and must be estimated from data. Nevertheless, this perspective provides us with a structured way to estimate them from data. 
An important issue is that Eq.~\ref{eqn:p} requires integration over the latent spaces, which is computationally cumbersome during both training and testing. To overcome this issue we lower bound Eq.~\ref{eqn:p} by a straightforward application of Jensen's inequality:
\begin{align}\label{eqn:lb}
& \hspace{-3mm} \log p( y^{(st)} | \mathbf{x}^{(s)}, \mathbf{x}^{(t)}) \\ 
& \geq \max_{\mathbf{z}^{(s)},\mathbf{z}^{(t)}}  \log p(\mathbf{z}^{(s)}|\mathbf{x}^{(s)})p(\mathbf{z}^{(t)}|\mathbf{x}^{(t)})p(y^{(st)}|\mathbf{z}^{(s)}, \mathbf{z}^{(t)}).\nonumber
\end{align}
In training and testing below, we employ this lower bound (\ie the right hand-side (RHS) in Eq.~\ref{eqn:lb}) as a surrogate for the exact but cumbersome similarity function between source and target domains. That is,
\begin{align}\label{eqn:gen_sim}
F(\mathbf{x}^{(s)}, \mathbf{x}^{(t)}, y^{(st)}) \triangleq & \max_{\mathbf{z}^{(s)},\mathbf{z}^{(t)}} \left\{ \log p(\mathbf{z}^{(s)}|\mathbf{x}^{(s)})\right. \nonumber \\
& \hspace{-20mm} \left. + \log p(\mathbf{z}^{(t)}|\mathbf{x}^{(t)}) + \log p(y^{(st)}|\mathbf{z}^{(s)}, \mathbf{z}^{(t)})\right\}.
\end{align}
Note that here $\log p(\mathbf{z}^{(s)}|\mathbf{x}^{(s)}), \log p(\mathbf{z}^{(t)}|\mathbf{x}^{(t)})$ are actually {\em data fitting} terms to restrict the feasible parameter spaces for $\mathbf{z}^{(s)}, \mathbf{z}^{(t)}$, respectively, performing the same functionality of regularization from the perspective of optimization. $\log p(y^{(st)}|\mathbf{z}^{(s)}, \mathbf{z}^{(t)})$ is essentially the {\em latent similarity measure} term in the joint latent space with embeddings. In the following section we show how many of the existing works in the literature can be viewed as special cases of our probabilistic framework.

\subsubsection{Relationship to Existing Works}\label{sssec:generalization}
Our probabilistic model can be considered as generalization of many embedding methods for ZSL. In particular, we will show that label embedding \cite{akata2013label}, output embedding \cite{Akata2015}, semantic similarity embedding \cite{Zhang2015}, deep neural network based embedding \cite{Ba2015}, and latent embedding \cite{Xian2016Latent} can all be viewed as special cases. For concreteness, we follow the notation in the original papers of each work and show how to view them as special cases of our model. 

%

{\em (\rmnum{1}) \underline{ Label embedding \cite{akata2013label}.}} This approach defines a bilinear prediction function as follows:
\begin{align}\label{eqn:dec}
f(x; \mathbf{W}) = \argmax_{y\in\mathcal{Y}} F(x,y;\mathbf{W}) = \argmax_{y\in\mathcal{Y}} \theta(x)^T\mathbf{W}\varphi(y),
\end{align}
where $F$ denotes the bilinear similarity function, $\theta(x), \varphi(y)$ denote the original image embedding and label embedding for image $x$ and label $y$, respectively. The matrix $\mathbf{W}$ is the parameter describing the bilinear classifier. In this work label embeddings are viewed as side information, for instance as attribute vectors.

We can view \cite{akata2013label} as a special case of our general probabilistic model as follows.  Define $\mathbf{x}^{(s)}\triangleq y, \mathbf{x}^{(t)}\triangleq x$. The three log-likelihoods in Eq. \ref{eqn:gen_sim} are described as follows:
\begin{align}
& \log p_B(\mathbf{z}^{(s)}|\mathbf{x}^{(s)}) = \left\{
\begin{array}{ll}
0, & \mbox{if} \, \mathbf{z}^{(s)} = \varphi(y) \\
-\infty, & \mbox{otherwise}
\end{array}
\right.\\
& \log p_D(\mathbf{z}^{(t)}|\mathbf{x}^{(t)}) = \left\{
\begin{array}{ll}
0, & \mbox{if} \, \mathbf{z}^{(t)} = \theta(x) \\
-\infty, & \mbox{otherwise}
\end{array}
\right.\\
& \log p_W(y^{(st)}| \mathbf{z}^{(s)}, \mathbf{z}^{(t)}) \triangleq  F(x,y;\mathbf{W}).
\end{align}
It can directly be verified by direct substitution that this is identical to the model described in \cite{akata2013label}.

{\em (\rmnum{2}) \underline{ Output embedding \cite{Akata2015}.}}  The similarity function proposed here is: 
\begin{align}
& F(x,y;\{\mathbf{W}\}_{1,\cdots,K}) = \sum_k \alpha_k\theta(x)^T\mathbf{W}_k\varphi_k(y), \\
& \mbox{s.t.} \, \sum_k\alpha_k=1, \nonumber
\end{align}
where $\{\mathbf{W}\}_{1,\cdots,K}$ denotes the parameters for $K$ different bilinear functions, $\varphi_k(y)$ denotes the $k$-th type of label embedding, and $\alpha_k$ denotes the combination weight for the $k$-th bilinear function. Then Eq. \ref{eqn:dec} with the above similarity function is utilized as the prediction function.

To view \cite{Akata2015} as a special case of our general probabilistic model, we can parametrize our model in the same way as we did for \cite{akata2013label}, except that 
\begin{align}
\hspace{-1mm}\log p_B(\mathbf{z}^{(s)}|\mathbf{x}^{(s)}) & = \sum_k \log p_B(\mathbf{z}_k^{(s)}|\varphi_k(y)) \nonumber \\ 
& \hspace{-5mm} = \left\{
\begin{array}{ll}
-\log K, & \mbox{if} \, \mathbf{z}_k^{(s)} = \varphi_k(y), \forall k, \\
-\infty, & \mbox{otherwise}
\end{array}
\right. 
\end{align}
\begin{align}
\log p_W(y^{(st)}| \mathbf{z}^{(s)}, \mathbf{z}^{(t)}) \triangleq F(x,y;\{\mathbf{W}\}_{1,\cdots,K}).
\end{align}
It can directly be verified by direct substitution that this is identical to the model described in \cite{Akata2015}.

{\em (\rmnum{3}) \underline{ Semantic similarity embedding \cite{Zhang2015}.}} Given a label embedding $\mathbf{c}$, \cite{Zhang2015} solves the following sparse coding problem to compute the semantic similarity embedding (SSE) for source domain:
\begin{equation}\label{eqn:simple}
\psi(\mathbf{c})=\argmin_{\boldsymbol{\alpha}\in \Delta^{|{\cal S}|}} \left\{\frac{\gamma}{2}\|\boldsymbol{\alpha}\|^2+\frac{1}{2}\|\mathbf{c}-\sum_{y\in\mathcal{S}}\mathbf{c}_y \alpha_y\|^2\right\},
\end{equation}
where $\gamma\geq 0$ is a predefined regularization parameter, $\|\cdot\|$ denotes the $\ell_2$ norm of a vector, and $\boldsymbol{\alpha} = (\alpha_y)_{y \in {\cal S}}$ describes contributions of different seen classes. 
Given a target-domain image embedding $\mathbf{x}$, the corresponding SSE is defined as 
\begin{align}
\phi_y(\mathbf{x})=\min(\mathbf{x}, \mathbf{v}_y),  \, \mbox{or} \; \phi_y(\mathbf{x})=\max(\mathbf{0}, \mathbf{x}-\mathbf{v}_y),
\end{align}
where $\mathbf{v}_y$ denotes a parameter for class $y$ that needs to be learned. Then the similarity function in \cite{Zhang2015} is defined as 
\begin{align}\label{eqn:Psi}
F(\mathbf{x},y;\mathbf{w})=\sum_{s\in\mathcal{S}}\left\langle\mathbf{w}, \phi_s(\mathbf{x})\right\rangle z_{y,s},
\end{align}
where $\mathcal{S}$ denotes the set of seen classes, $z_{y,s}$ denotes the $s$-th entry in the SSE for class $y$, and $\mathbf{w}$ denotes the classifier parameter. Then Eq. \ref{eqn:dec} with the above similarity function is utilized as the prediction function.

To view \cite{Zhang2015} as a special case of our general probabilistic model, we can use the same methodology to model the three log-likelihoods in Eq. \ref{eqn:gen_sim} as follows:
\begin{align}
& \log p_B(\mathbf{z}^{(s)}|\mathbf{x}^{(s)}) = \left\{
\begin{array}{ll}
0, & \mbox{if} \, \mathbf{z}^{(s)} = \psi(\mathbf{x}^{(s)}) \\
-\infty, & \mbox{otherwise}
\end{array}
\right.\\
& \log p_D(\mathbf{z}^{(t)}|\mathbf{x}^{(t)}) = \left\{
\begin{array}{ll}
0, & \mbox{if} \, \mathbf{z}^{(t)} = \phi(\mathbf{x}^{(t)}) \\
-\infty, & \mbox{otherwise}
\end{array}
\right.\\
& \log p_W(y^{(st)}| \mathbf{z}^{(s)}, \mathbf{z}^{(t)}) \triangleq F(\mathbf{x},y;\mathbf{w}).
\end{align}


{\em (\rmnum{4}) \underline{ Deep neural network based embedding \cite{Ba2015}.}} The prediction function in \cite{Ba2015} is the same as Eq. \ref{eqn:dec}, except that now functions $\varphi, \theta$ are learned using neural networks, and the learned $\mathbf{W}$ represents the weight for a fully-connected layer between the two embeddings from source and target domains, respectively.. Therefore, in test time we can use the same parametrization for our model so that \cite{Ba2015} can be taken as our special case mathematically.

{\em (\rmnum{5}) \underline{ Latent embedding \cite{Xian2016Latent}}}. This approach learns the latent embedding spaces explicitly based on clustering. For each cluster a bilinear classifier is learned for measuring similarities. Correspondingly the similarity decision function in \cite{Xian2016Latent} is defined as follows:
\begin{align}\label{eqn:lat_dec}
F(\mathbf{x}, \mathbf{y}; \{\mathbf{W}\}_{1,\cdots,K}) = \max_{1\leq i\leq K} \mathbf{x}^T\mathbf{W}_i\mathbf{y},
\end{align}
where $\mathbf{x}, \mathbf{y}$ denote image and label embeddings, respectively, and $i$ denotes the $i$-th bilinear classifier with parameter $\mathbf{W}_i$ among the $K$ classifiers. Because of the $\max$ operator, the indicator variable $i$ becomes the latent variable for selecting which bilinear classifier should be utilized per data pair.

\begin{figure}[t]
	\centerline{\includegraphics[width=\linewidth]{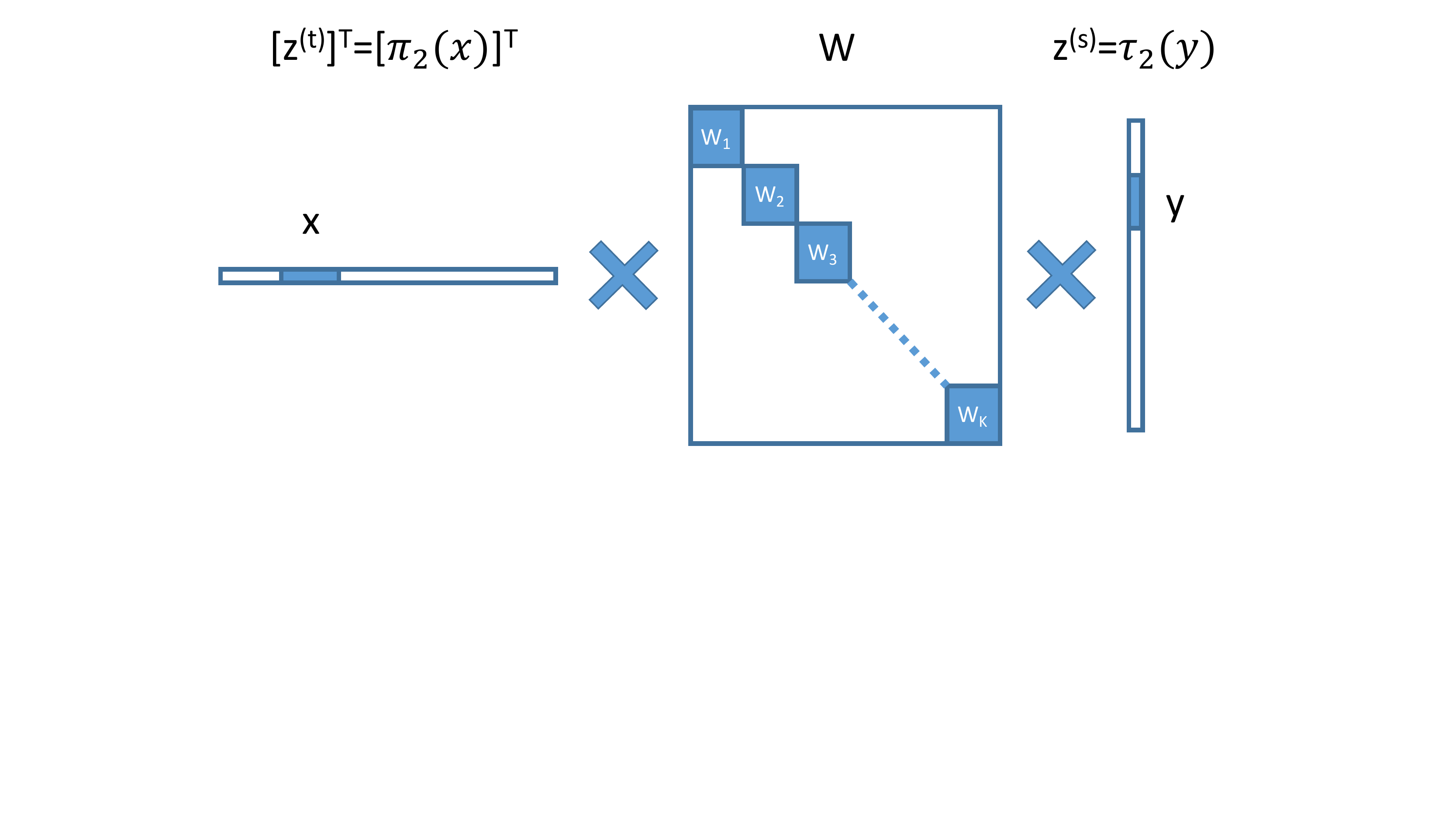}}
	\vspace{-18mm}
	\caption{\footnotesize{Illustration of our generalization for latent embedding \cite{Xian2016Latent}. This figure illustrates the similarity measure of $\mathbf{x}^T\mathbf{W}_2\mathbf{y}$. By searching for the maximum over different combinations of $\mathbf{z}^{(t)}, \mathbf{z}^{(s)}$, our model can exactly represent the prediction function in Eq. \ref{eqn:lat_dec}.}}\label{fig:latent_gen}
	\vspace{-3mm}
\end{figure}

To view \cite{Xian2016Latent} as a special case of our general probabilistic model, we first construct a new $\mathbf{W}$ in Eq. \ref{eqn:dec} by putting $\mathbf{W}_i, \forall i$ as {\em blocks} along the diagonal, \ie $\mathbf{W} \triangleq diag(\mathbf{W}_1, \cdots, \mathbf{W}_K)\in\mathbb{R}^{Kd_t\times Kd_s}$, where $d_t, d_s$ denote the dimensions of $\mathbf{x}, \mathbf{y}$ in Eq. \ref{eqn:lat_dec}, respectively, and filling in the rest entries with zeros. Here, along either columns or rows in $\mathbf{W}$ there exist $K$ blocks with dimensionality of either $d_t$ or $d_s$ per block. Then we design two functions $\pi: \mathbb{R}^{d_t}\rightarrow\mathbb{R}^{Kd_t}, \tau: \mathbb{R}^{d_s}\rightarrow\mathbb{R}^{Kd_s}$ to map the original data $\mathbf{x}, \mathbf{y}$ to higher dimensional spaces with $K$ blocks, respectively. The functionality of $\pi, \tau$ is to assign $\mathbf{x}, \mathbf{y}$ to one block $i, j\in[K]$, denoted by $\pi_i(\mathbf{x}), \tau_j(\mathbf{y})$, and fill in the rest entries using zeros. The whole construction procedure is illustrated in Fig. \ref{fig:latent_gen}.
Now we can use the same methodology to model the three log-likelihoods in Eq. \ref{eqn:gen_sim} as follows:
\begin{align}
& \log p_B(\mathbf{z}^{(s)}|\mathbf{x}^{(s)}) = \left\{
\begin{array}{ll}
-\log K, & \mbox{if} \, \mathbf{z}^{(s)} = \tau_j(\mathbf{y}), \forall j \\
-\infty, & \mbox{otherwise}
\end{array}
\right.\\
& \log p_D(\mathbf{z}^{(t)}|\mathbf{x}^{(t)}) = \left\{
\begin{array}{ll}
-\log K, & \mbox{if} \, \mathbf{z}^{(t)} = \pi_i(\mathbf{x}), \forall i \\
-\infty, & \mbox{otherwise}
\end{array}
\right.\\
& \log p_W(y^{(st)}| \mathbf{z}^{(s)}, \mathbf{z}^{(t)}) \triangleq \left[\mathbf{z}^{(t)}\right]^T\mathbf{W}\mathbf{z}^{(s)} + \Delta(i, j),
\end{align}
where $\Delta(i,j)=0$ if $i=j$, otherwise $-\infty$, which enforces $\pi, \tau$ to select the same block for similarity measure.

In the light of these observations we can view our framework as a way to describe different modes of data in a unified semantic space. Central to this observation is the key insight that zero-shot learning is fundamentally about detecting whether or not an arbitrary tuple $(\mathbf{x}^{(s)},\mathbf{x}^{(t)})$ is associated with the same underlying label or not. This question is then fundamentally about binary classification. A second aspect of our framework is the latent embedding. This latent embedding describes source and target domain realizations as being conditionally independent of each other given their latent embeddings. For instance, this enforces the natural assumption that an image is conditionally independent of its textual description if it is conditioned on visual attributes, which serve as the latent embedding. In this way latent embeddings serve as sufficient statistics for identifying similarity of the tuple. This perspective in turn serves to unify many of the existing works in the literature. Nevertheless, for the concreteness we must choose specific likelihood functions. We propose a joint supervised dictionary learning approach in Sec. \ref{ssec:sdl}.



\subsubsection{Training}\label{sssec:training}
During training time, we are given independent source and target domain instances, $\mathbf{x}^{(s)}_i,\mathbf{x}^{(t)}_j$, and a binary label $y_{ij}^{(st)}$ indicating whether or not they belong to the same class. We parametrize the probability kernels in Eq. \ref{eqn:lb} using $p_B(\mathbf{z}^{(s)}|\mathbf{x}^{(s)})$, $p_D(\mathbf{z}^{(t)}|\mathbf{x}^{(t)})$, $p_W(y^{(st)}|\mathbf{z}^{(s)}, \mathbf{z}^{(t)})$ in terms of {\em data-independent} parameters $B,\,D,\,W$ respectively, and estimate them discriminatively using training data.

\begin{algorithm}[t]\footnotesize
	\SetAlgoLined
	\SetKwInOut{Input}{Input}\SetKwInOut{Output}{Output}
	\Input{training data $\{(\mathbf{x}_i^{(s)}, y_i^{(s)})\}$ and $\{(\mathbf{x}_j^{(t)}, y_j^{(t)})\}$}
	\Output{$B,D,W$}
	\BlankLine
	Initialize $B, D$; \\
	$\forall i, \mathbf{z}_i^{(s)}\leftarrow\argmax_{\mathbf{z}^{(s)}}\log p_B(\mathbf{z}^{(s)}|\mathbf{x}^{(s)}_i)$; \\
	$\forall j, \mathbf{z}_j^{(t)}\leftarrow\argmax_{\mathbf{z}^{(t)}}\log p_D(\mathbf{z}^{(t)}|\mathbf{x}^{(t)}_j)$; \\
	$W\leftarrow\argmax_W\sum_{i=1}^C\sum_{j=1}^N\log p_W(y^{(st)}_{ij} | \mathbf{z}^{(s)}_i, \mathbf{z}^{(t)}_j)$;\\
	\Repeat{Converge to a local maximum}
	{ 
		\ForEach{$i$}{
			\ForEach{$j$}{
				$\mathbf{z}_{ij}^{(s)}\leftarrow\mathbf{z}_{i}^{(s)}$;
				$\mathbf{z}_{ij}^{(t)}\leftarrow\mathbf{z}_{j}^{(t)}$;\\
				\Repeat{Converge to a local maximum}
				{
					$\mathbf{z}_{ij}^{(s)}\leftarrow\argmax_{\mathbf{z}^{(s)}}\log p_B(\mathbf{z}^{(s)}|\mathbf{x}^{(s)}_i) + \log p_W(y^{(st)}_{ij} | \mathbf{z}^{(s)}, \mathbf{z}_{ij}^{(t)})$; \\
					$\mathbf{z}_{ij}^{(t)}\leftarrow\argmax_{\mathbf{z}^{(t)}}\log p_D(\mathbf{z}^{(t)}|\mathbf{x}^{(t)}_j) + \log p_W(y^{(st)}_{ij} | \mathbf{z}_{ij}^{(s)}, \mathbf{z}^{(t)})$; 					
				}
			}
		}		
		$B\leftarrow\argmax_B\sum_{i=1}^C\sum_{j=1}^{N} \log p_B(\mathbf{z}_{ij}^{(s)}|\mathbf{x}^{(s)}_i)$;\\
		$D\leftarrow\argmax_D\sum_{i=1}^{C}\sum_{j=1}^N \log p_D(\mathbf{z}_{ij}^{(t)}|\mathbf{x}^{(t)}_j)$;\\
		$W\leftarrow\argmax_W\sum_{i=1}^C\sum_{j=1}^N\log p_W(y^{(st)}_{ij} | \mathbf{z}_{ij}^{(s)}, \mathbf{z}_{ij}^{(t)})$;
	}
	\Return $B,D,W$
	\caption{Jointly latent embedding learning algorithm for solving Eq. \ref{eqn:gen_obj}}\label{alg:gen_train}
\end{algorithm}

Note that maximizing the RHS in Eq.~\ref{eqn:lb} over latent embeddings $\mathbf{z}^{(s)}, \mathbf{z}^{(t)}$ is actually a joint optimization which needs to be conducted for every pair of source and target data instances $(\mathbf{x}^{(s)}, \mathbf{x}^{(t)})$. Therefore, in order to maximize the lower bound of the log-likelihood over the entire training data, we propose the following joint optimization problem for learning the parameters $B, D, W$:
\begin{align}\label{eqn:gen_obj}
\max_{B,D,W} & \sum_{i=1}^C\sum_{j=1}^N \max_{\mathbf{z}^{(s)},\mathbf{z}^{(t)}}\left\{  \log p_B(\mathbf{z}^{(s)}|\mathbf{x}^{(s)}_i) \right. \\
& \left. + \log p_D(\mathbf{z}^{(t)}|\mathbf{x}^{(t)}_j) + \log p_W(y^{(st)}_{ij} | \mathbf{z}^{(s)}, \mathbf{z}^{(t)}) \right\}, \nonumber
\end{align}
where $C$ is the size of the source domain training data (\ie number of observed class labels) and $N$ is the size of the target domain training data.

Here we emphasize the fact that any pair of latent embeddings $\left(\mathbf{z}^{(s)}, \mathbf{z}^{(t)}\right)$ in Eq. \ref{eqn:gen_obj} are essentially {\em fully coupled}, \ie one is a function of the other.  In other words, the source (\resp target) domain latent embeddings should change with different target (\resp source) domain latent embeddings. This naturally suggests an alternating optimization mechanism for solving Eq. \ref{eqn:gen_obj} as shown in Alg. \ref{alg:gen_train}. However, as we see clearly, this algorithm would lead to significantly high computational complexity because of the optimization for every pair of latent embeddings in source and target domains, especially for large-scale data. 

Instead as a compromise for running speed, we propose the following training objective as the lower bound of Eq.~\ref{eqn:gen_obj} over the source and target domain data by pulling the operator $\max_{\mathbf{z}^{(s)}, \mathbf{z}^{(t)}}$ out of double-summations:
\begin{align}\label{eqn:obj}
& \max_{B,D,W} \max_{\{\mathbf{z}^{(s)}_i\},\{\mathbf{z}^{(t)}_j\}} N\sum_{i=1}^C \log p_B(\mathbf{z}^{(s)}_i|\mathbf{x}^{(s)}_i) \\
& + C\sum_{j=1}^N \log p_D(\mathbf{z}^{(t)}_j|\mathbf{x}^{(t)}_j) + \sum_{i=1}^C\sum_{j=1}^N\log p_W(y^{(st)}_{ij} | \mathbf{z}^{(s)}_i, \mathbf{z}^{(t)}_j). \nonumber
\end{align}
Although in this relaxation $\mathbf{z}^{(s)}, \mathbf{z}^{(t)}$ are still coupled, the latent embeddings for both source and target domain data are fixed. That is, for $\mathbf{x}_i^{(s)}, \forall i$ (\resp $\mathbf{x}_j^{(t)}, \forall j$), there exists only one corresponding latent embedding $\mathbf{z}_i^{(s)}$ (\resp $\mathbf{z}_j^{(t)}$). Therefore, fundamentally different from Eq. \ref{eqn:gen_obj}, the relaxation in Eq. \ref{eqn:obj} significantly reduces the computational complexity of our model in training time. In the rest of paper, we consider Eq. \ref{eqn:obj} as our training objective by default without explicit mention.\\ 

\noindent
{\bf Salient Aspects of our Training Algorithm:} Based on Eq.~\ref{eqn:obj} our objective is two-fold. We need to learn a low-dimensional latent embedding that not only accurately represents the observed data in each domain but also is capable of inferring cross-domain statistical relationships when one exists. Note that the first two log-likelihoods in Eq.~\ref{eqn:obj} are data fitting terms, and the last one measures the {\em joint latent similarity} between the two latent vectors.

With this insight we propose a general alternating optimization algorithm to jointly learn $\{\mathbf{z}_i^{(s)}\}, \{\mathbf{z}_j^{(t)}\}, B, D, W$ in Eq. \ref{eqn:obj} in Alg. \ref{alg:train}. This follows from the exchangeability of two $\max$ operators. In this way our learning algorithm guarantees {\em convergence} to a local optimum within finite number of iterations. Also since the update rules for $\forall i, \mathbf{z}_i^{(s)}$ (or $\forall j, \mathbf{z}_j^{(t)}$) are independent given $\forall j, \mathbf{z}_j^{(t)}$ (or $\forall i, \mathbf{z}_i^{(s)}$) and parameters $B,D,W$, we can potentially utilize parallel or distributed computing to train our models. This has obvious computational benefits.

Our approach diverts from some of the previous works such as \cite{hariharan2012efficient} where source domain vectors for unseen classes are also known during training. This perspective lets one exploit knowledge of unseen source domain classes during training. In contrast we are not provided unseen data for either the source or target domains. Thus, our data-independent variables $B,D,W$ do not contain any information about unseen data.

\begin{algorithm}[t]\footnotesize
	\SetAlgoLined
	\SetKwInOut{Input}{Input}\SetKwInOut{Output}{Output}
	\Input{training data $\{(\mathbf{x}_i^{(s)}, y_i^{(s)})\}$ and $\{(\mathbf{x}_j^{(t)}, y_j^{(t)})\}$}
	\Output{$\{\mathbf{z}_i^{(s)}\}, \{\mathbf{z}_j^{(t)}\},B,D,W$}
	\BlankLine
	Initialize $B, D$; \\
	$\forall i, \mathbf{z}_i^{(s)}\leftarrow\argmax_{\mathbf{z}^{(s)}}\log p_B(\mathbf{z}^{(s)}|\mathbf{x}^{(s)}_i)$; \\
	$\forall j, \mathbf{z}_j^{(t)}\leftarrow\argmax_{\mathbf{z}^{(t)}}\log p_D(\mathbf{z}^{(t)}|\mathbf{x}^{(t)}_j)$; \\
	$W\leftarrow\argmax_W\sum_{i=1}^C\sum_{j=1}^N\log p_W(y^{(st)}_{ij} | \mathbf{z}^{(s)}_i, \mathbf{z}^{(t)}_j)$;\\
	\Repeat{Converge to a local maximum}
	{ 
		$\forall i, \mathbf{z}_i^{(s)}\leftarrow\argmax_{\mathbf{z}^{(s)}}\log p_B(\mathbf{z}^{(s)}|\mathbf{x}^{(s)}_i) + \sum_{j=1}^N\log p_W(y^{(st)}_{ij} | \mathbf{z}^{(s)}, \mathbf{z}^{(t)}_j)$; \\
		$\forall j, \mathbf{z}_j^{(t)}\leftarrow\argmax_{\mathbf{z}^{(t)}}\log p_D(\mathbf{z}^{(t)}|\mathbf{x}^{(t)}_j) + \sum_{i=1}^C\log p_W(y^{(st)}_{ij} | \mathbf{z}^{(s)}_i, \mathbf{z}^{(t)})$; \\
		$B\leftarrow\argmax\sum_{i=1}^C \log p_B(\mathbf{z}^{(s)}_i|\mathbf{x}^{(s)}_i)$;\\
		$D\leftarrow\argmax\sum_{j=1}^N \log p_D(\mathbf{z}^{(t)}_j|\mathbf{x}^{(t)}_j)$;\\
		$W\leftarrow\argmax_W\sum_{i=1}^C\sum_{j=1}^N\log p_W(y^{(st)}_{ij} | \mathbf{z}^{(s)}_i, \mathbf{z}^{(t)}_j)$;
	}
	\Return $\{\mathbf{z}_i^{(s)}\}, \{\mathbf{z}_j^{(t)}\}, B,D,W$
	\caption{Simplified jointly latent embedding learning algorithm for solving Eq. \ref{eqn:obj}}\label{alg:train}
\end{algorithm}

\subsubsection{Testing}
In order to avoid confusion we index unseen class data with $i',\,j'$ corresponding to source and target domain respectively. The seen class training data is indexed as before with $i,j$. During test time the source domain data $\{({\mathbf{x}}_{i'}^{(s)}, {y}_{i'}^{(s)})\}$ for all the unseen classes are revealed. We are then presented with an instance of unseen target domain data, $\{{\mathbf{x}}_{j'}^{(t)}\}$. Our objective is to identify an unseen source domain vector that best matches the unseen instance.

Considering Eq. \ref{eqn:gen_sim} and Eq. \ref{eqn:gen_obj}, naturally we have the following test-time decision function:
\begin{align}\label{eqn:gen_dec}
& {y}_{j'}^{(t)} = {y}_{i'_*}^{(s)}, \; \mbox{s.t.} \, i'_*=\argmax_{i'\in[C']}\Big\{\max_{\mathbf{z}^{(s)},\mathbf{z}^{(t)}}\left\{  \log p_B(\mathbf{z}^{(s)}|\mathbf{x}^{(s)}_{i'}) \right. \nonumber \\
& \left. + \log p_D(\mathbf{z}^{(t)}|\mathbf{x}^{(t)}_{j'}) + \log p_W(y_{i'j'}^{(st)}=1| \mathbf{z}^{(s)}, \mathbf{z}^{(t)}) \right\}\Big\}, 
\end{align}
where $C'$ and $[C']$ denote the number of unseen classes and the index set of unseen classes starting from 1, respectively.

\begin{algorithm}[t]\footnotesize
	\SetAlgoLined
	\SetKwInOut{Input}{Input}\SetKwInOut{Output}{Output}
	\Input{test data $\{({\mathbf{x}}_{i'}^{(s)}, {y}_{i'}^{(s)})\}$ and $\{{\mathbf{x}}_{j'}^{(t)}\}$; learned parameters $B,D,W$ during training
	}
	\Output{$\{y_{j'}^{(t)}\}$}
	\BlankLine
	$\forall i', {\mathbf{z}}_{i'}^{(s)}\leftarrow\argmax_{{\mathbf{z}}_{i'}^{(s)}}\log p_B({\mathbf{z}}^{(s)}_{i'}|{\mathbf{x}}^{(s)}_{i'})$; \\
	$\forall j', {\mathbf{z}}_{j'}^{(t)}\leftarrow\argmax_{{\mathbf{z}}_{j'}^{(t)}}\log p_D({\mathbf{z}}^{(t)}_{j'}|{\mathbf{x}}^{(t)}_{j'})$;\\
	\ForEach{$j'$}{
		$\mathcal{S}\leftarrow\emptyset$;\\
		\ForEach{$i'$}{
			$\mathbf{z}_{i'j'}^{(s)}\leftarrow\mathbf{z}_{i'}^{(s)}$;
			$\mathbf{z}_{i'j'}^{(t)}\leftarrow\mathbf{z}_{j'}^{(t)}$;\\
			\Repeat{Converge to a local maximum}{
				$\mathbf{z}_{i'j'}^{(s)}\leftarrow\argmax_{\mathbf{z}^{(s)}}\log p_B(\mathbf{z}^{(s)}|\mathbf{x}^{(s)}_{i'}) + \log p_W(y^{(st)}_{i'j'} | \mathbf{z}^{(s)}, \mathbf{z}_{i'j'}^{(t)})$; \\
				$\mathbf{z}_{i'j'}^{(t)}\leftarrow\argmax_{\mathbf{z}^{(t)}}\log p_D(\mathbf{z}^{(t)}|\mathbf{x}^{(t)}_{j'}) + \log p_W(y^{(st)}_{i'j'} | \mathbf{z}_{i'j'}^{(s)}, \mathbf{z}^{(t)})$; 		
			}
			$\mathcal{S}\leftarrow[\mathcal{S}; \log p_B(\mathbf{z}_{i'j'}^{(s)}|\mathbf{x}^{(s)}_{i'}) + \log p_D(\mathbf{z}_{i'j'}^{(t)}|\mathbf{x}^{(t)}_{j'}) + \log p_W(y^{(st)}_{i'j'} | \mathbf{z}_{i'j'}^{(s)}, \mathbf{z}_{i'j'}^{(t)})$;
		}
		$[s, i'_*] \leftarrow \max(\mathcal{S})$; 
		$y_{j'}^{(t)} \leftarrow y_{i'_*}^{(s)}$;
	}		
	\Return $\{y_{j'}^{(t)}\}$
	\caption{Joint latent embedding testing algorithm}\label{alg:gen_test}
\end{algorithm}

Similar to solving Eq. \ref{eqn:gen_obj} in training time, Eq. \ref{eqn:gen_dec} also suggests an alternating optimization algorithm to determine the maximum similarity between any pair of unseen source and target domain data, as shown in Alg. \ref{alg:gen_test}. Still the high computational complexity here prevents it from being used for large-scale data.

Alternatively we adopt the strategy in the relaxation of Eq. \ref{eqn:dec} to reduce the test-time computational complexity. That is, we would like to estimate the fixed latent embeddings for all the unseen source and target domain data so that prediction of the unseen classes is deterministic. In this way, there will be no $\max_{\mathbf{z}^{(s)}, \mathbf{z}^{(t)}}$ involved in Eq. \ref{eqn:gen_dec}. To better estimate such embeddings we are also given seen class latent embeddings $\{\mathbf{z}_i^{(s)}\}$ and $\{\mathbf{z}_j^{(t)}\}$ and the parameters $B,D,W$ that are all learned during training. 
This naturally suggests the optimization algorithm in Alg.~\ref{alg:test} by adapting the training algorithm in Alg. \ref{alg:train} to test time scenarios. Note that while the second term during this estimation process appears unusual we are merely exploiting the fact that the unseen class has no intersection with seen classes. Consequently, we can assume that $y_{i'j}^{(st)}=-1, \,\,y_{ij'}^{(st)}=-1$. Notice that the latent vector computation is again amenable to fast parallel or distributed computing.  

\begin{algorithm}[t]\footnotesize
	\SetAlgoLined
	\SetKwInOut{Input}{Input}\SetKwInOut{Output}{Output}
	\Input{test data $\{({\mathbf{x}}_{i'}^{(s)}, {y}_{i'}^{(s)})\}$ and $\{{\mathbf{x}}_{j'}^{(t)}\}$; learned latent embeddings for seen classes (training data) $\{\mathbf{z}_i^{(s)}\}$ and $\{\mathbf{z}_j^{(t)}\}$; learned parameters $B,D,W$ during training
	}
	\Output{$\{{\mathbf{z}}_{i'}^{(s)}\}, \{{\mathbf{z}}_{j'}^{(t)}\}$}
	\BlankLine
	$\forall i', {\mathbf{z}}_{i'}^{(s)}\leftarrow\argmax_{{\mathbf{z}}_{i'}^{(s)}}\log p_B({\mathbf{z}}^{(s)}_{i'}|{\mathbf{x}}^{(s)}_{i'}) + \sum_{j=1}^N\log p_W(-1|{\mathbf{z}}^{(s)}_{i'}, \mathbf{z}^{(t)}_j)$; \\
	$\forall j', {\mathbf{z}}_{j'}^{(t)}\leftarrow\argmax_{{\mathbf{z}}_{j'}^{(t)}}\log p_D({\mathbf{z}}^{(t)}_{j'}|{\mathbf{x}}^{(t)}_{j'}) + \sum_{i=1}^C\log p_W(-1|\mathbf{z}^{(s)}_i, {\mathbf{z}}^{(t)}_{j'})$;\\
	\Return $\{{\mathbf{z}}_{i'}^{(s)}\}, \{{\mathbf{z}}_{j'}^{(t)}\}$\;
	\caption{Test-time estimation of latent embeddings}\label{alg:test}
\end{algorithm}


\noindent\\
{\bf Decision function:} We next compute the likelihood of being the same class label, \ie $p({y}_{i'j'}^{(st)}=1|{\mathbf{x}}_{i'}^{(s)}, {\mathbf{x}}_{j'}^{(t)})$, for an arbitrary target domain data ${\mathbf{x}}_{j'}^{(t)}$ using the source domain data $({\mathbf{x}}_{i'}^{(s)}, {y}_{i'}^{(s)})$. 
Based on Eq. \ref{eqn:gen_dec} there are two options: The first option is to directly employ latent estimates $\mathbf{z}^{(s)}_{i'}, \mathbf{z}_{j'}^{(t)}$ for $\mathbf{x}^{(s)}_{i'}, \mathbf{x}_{j'}^{(t)}$, respectively, and ignore the two data fitting terms. This leads to the following expression (which is evidently related to the one employed in \cite{akata2013label, Ba2015, Zhang2015}):
\begin{align}\label{eqn:decision1}
{y}_{j'}^{(t)} ={y}_{i'_*}^{(s)}, \mbox{s.t.} i'_*=\argmax_{i'}\left\{\log p_W({y}_{i'j'}^{(st)}=1|{\mathbf{z}}^{(s)}_{i'}, {\mathbf{z}}_{j'}^{(t)})\right\}.
\end{align}
A second option is to use Eq. \ref{eqn:gen_dec} with fixed $\mathbf{z}_{i'}^{(s)}, \mathbf{z}_{j'}^{(t)}$ for prediction, which in turn leads us to:
\begin{align}\label{eqn:decision2}
{y}_{j'}^{(t)} = {y}_{i'_*}^{(s)}, \mbox{s.t.} i'_* & = \argmax_{i'}\Big\{ \log p_B({\mathbf{z}}^{(s)}_{i'}|{\mathbf{x}}^{(s)}_{i'}) \nonumber \\
& + \log p_W({y}_{i'j'}^{(st)}=1|{\mathbf{z}}^{(s)}_{i'}, {\mathbf{z}}_{j'}^{(t)})\Big\}. 
\end{align}


Note that the decision function in Eq. \ref{eqn:decision2} is different from the one in Eq. \ref{eqn:decision1}, which is widely used in embedding methods (see Sec. \ref{sssec:generalization}). In Eq. \ref{eqn:decision2} we also penalize source domain fit to identify the class label. Intuitively this choice optimizes the source domain embedding that best aligns with the target data. One reason for doing so is based on the fact that our information is asymmetric and the single source domain vector per class represents the strongest information about the class. Therefore, our attempt is to penalize the loss functions towards a source domain fit. 

In general one could also view source domain embeddings $\mathbf{z}^{(s)}$ as a parameter in Eq.~\ref{eqn:decision2} and optimize it as well. This is computationally somewhat more expensive. While more experiments maybe necessary to see whether or not this leads to improved performance, we have not found this additional degree of freedom to significantly improve performance.


\subsection{Parametrization}\label{ssec:sdl}
In this section we develop a {\em supervised dictionary learning (SDL)} formulation to parametrize Eq.~\ref{eqn:obj}. 
Specifically, we map data instances into the latent space as the coefficients based on a learned dictionary, and formulate an empirical risk function as the similarity measure which attempts to minimize the regularized hinge loss with the joint latent embeddings.

For purpose of exposition we overload notation in Sec.~\ref{sssec:training} and let $\mathbf{B}\in\mathbb{R}^{d_s\times h_s},\mathbf{D}\in\mathbb{R}^{d_t\times h_t},\mathbf{W}\in\mathbb{R}^{h_s\times h_t}$ as the source domain dictionary, target domain dictionary, and the cross-domain similarity matrix in the joint latent space, respectively. Here $d_s$ and $d_t$ are original feature dimensions, and $h_s$ and $h_t$ are the sizes of dictionaries. Then given the seen class source domain data $\{(\mathbf{x}_i^{(s)}, y_i^{(s)})\}$ and target domain data $\{(\mathbf{x}_j^{(t)}, y_j^{(t)})\}$, we choose to parametrize the three log-likelihoods in Eq. \ref{eqn:obj}, denoted by $\log p_B, \log p_D, \log p_W$, respectively using dictionary learning and regularized hinge loss as follows. For source domain embedding, following \cite{Zhang2015}, we enforce source domain latent coefficients to lie on a simplex (see Eq. \ref{eqn:pb} below). 
For target domain embedding, we follow the convention. We allow the latent vectors to be arbitrary while constraining the elements in the dictionary to be within the unit ball. 
Specifically, $\forall i, \forall j$, we have,
\begin{align}\label{eqn:pb}
-\log p_B \triangleq \,\, & \, \frac{\lambda_1^{(s)}}{2}\|\mathbf{z}_i^{(s)}\|_2^2 + \frac{\lambda_2^{(s)}}{2}\|\mathbf{x}_i^{(s)} - \mathbf{B}\mathbf{z}_i^{(s)}\|_2^2, \\
& \mbox{s.t.} \quad \mathbf{z}_i^{(s)}\geq\mathbf{0}, \; \mathbf{e}^T\mathbf{z}_i^{(s)}=1, \nonumber \\
\label{eqn:pd}
-\log p_D \triangleq \,\, & \, \frac{\lambda_1^{(t)}}{2}\|\mathbf{z}_j^{(t)}\|_2^2 + \frac{\lambda_2^{(t)}}{2}\|\mathbf{x}_j^{(t)} - \mathbf{D}\mathbf{z}_j^{(t)}\|_2^2, \\
& \mbox{s.t.} \quad \forall k, \; \|\mathbf{D}_k\|_2^2\leq 1, \nonumber \\
\label{eqn:pw}
-\log p_W \triangleq \,\, & \, \frac{\lambda_W}{2}\|\mathbf{W}\|_F^2 + \Big\lfloor 1-\mathbf{1}_{y_{ij}^{(st)}}\left[\mathbf{z}_i^{(s)}\right]^T\mathbf{W}\mathbf{z}_j^{(t)}\Big\rfloor_+,
\end{align}
where $\|\cdot\|_F$ and $\|\cdot\|_2$ are the Frobenius norm and $\ell_2$ norm operators, $\lfloor\cdot\rfloor_+=\max\{0, \cdot\}$, $\geq$ is an entry-wise operator, $[\cdot]^T$ is the matrix transpose operator, $\mathbf{e}$ is a vector of 1's, and $\forall k, \mathbf{D}_k$ denotes the $k$-th row in the matrix $\mathbf{D}$. $\mathbf{1}_{y_{ij}^{(st)}}=1$ if $y_{i}^{(s)}=y_{j}^{(t)}$ and $-1$ otherwise. The regularization parameters $\lambda_1^{(s)}\geq0, \lambda_2^{(s)}\geq0, \lambda_1^{(t)}\geq0, \lambda_2^{(t)}\geq0, \lambda_W\geq0$ are fixed during training. Cross validation is used to estimate these parameters by holding out a portion of seen classes (see Sec.~\ref{ssec:implementation}). With sufficient data (\ie no need of regularization to avoid overfitting), our SDL approach indeed is equivalent to the relaxation of the following joint optimization problem:
\begin{align}
\hspace{-5mm} \min_{\substack{\{\mathbf{z}_i^{(s)}\}, \{\mathbf{z}_j^{(t)}\}, \\ \mathbf{W}, \mathbf{B}, \mathbf{D}}} & \sum_{i,j} \max\left\{0, 1-\mathbf{1}_{y_{ij}^{(st)}}\left[\mathbf{z}_i^{(s)}\right]^T\mathbf{W}\mathbf{z}_j^{(t)}\right\} \\
\mbox{s.t.} \hspace{5mm} & \mathbf{x}_i^{(s)} = \mathbf{B}\mathbf{z}_i^{(s)}, \, \mathbf{z}_i^{(s)}\geq0, \, \mathbf{e}^T\mathbf{z}_i^{(s)} = 1, \, \forall i, \nonumber \\
& \mathbf{x}_j^{(t)} = \mathbf{D}\mathbf{z}_j^{(t)}, \, \forall j, \, \|\mathbf{D}_k\|_2^2\leq 1, \, \forall k. \nonumber
\end{align}

Observe that our method leverages association between the source domain and target domain vectors across all seen classes and learns a single matrix for all classes. Our objective function utilizes a hinge loss to penalize mis-associations between source and target pairs in the joint latent space. 

\noindent
{\bf Training \& Cross-Validation:} We hold-out data corresponding to two randomly sampled seen classes and train our method using Alg. \ref{alg:train} on the rest of the seen classes for different combinations of regularization parameters. Training is performed by substituting Eq. \ref{eqn:pb}, \ref{eqn:pd}, and \ref{eqn:pw} into Alg.~\ref{alg:train}. For efficient computation, we utilize proximal gradient algorithms \cite{parikh2014proximal} with simplex projection \cite{duchi2008efficient} for updating $ \mathbf{z}_i^{(s)}, \forall i$ and $\mathbf{z}_j^{(t)}, \forall j$, respectively. We use linear SVMs to learn $\mathbf{W}$. 

\noindent
{\bf Testing:} We substitute Eq. \ref{eqn:pb}, \ref{eqn:pd}, and \ref{eqn:pw} into Alg.~\ref{alg:test} and run it by fixing all the parameters learned during training. This leads to estimation of the latent embeddings for unseen class source and target domain data. Then we apply Eq. \ref{eqn:decision1} or \ref{eqn:decision2} to predict the class label for target domain data.

\begin{table}[t]\footnotesize
\centering
\setlength\tabcolsep{3pt}
\caption{\footnotesize{Statistics of different datasets, where ``bin.'' and ``cont.'' stand for binary value and continuous value, respectively.}}\label{tab:dataset}\vspace{1mm}
\begin{tabular}{|l|lll|}
\hline
 Dataset & \# instances & \# attributes & \# seen/unseen classes \\
 \hline\hline  
aP\&Y & 15,339 & 64 (cont.) & 20 / 12 \\
AwA & 30,475 & 85 (cont.) & 40 / 10 \\
CUB-200-2011 & 11,788 & 312 (bin.) & 150 / 50 \\
SUN Attribute & 14,340 & 102 (bin.) & 707 / 10 \\
 \hline
\end{tabular}
\vspace{-4mm}
\end{table}

\begin{table*}[t]\footnotesize
\begin{minipage}{\linewidth}
\centering
\caption{\footnotesize{Zero-shot recognition accuracy comparison (\%) on the four datasets. Except for \cite{Akata2015} where AlexNet \cite{NIPS2012_4824} is utilized for extracting CNN features, for all the other methods we use vgg-verydeep-19 \cite{simonyan2014very} CNN features.}}\vspace{1mm}\label{tab:benchmark}
\begin{tabular}{|lllll|l|}
\hline
Method & aP\&Y & AwA & CUB-200-2011 & SUN Attribute & Ave.\\
\hline\hline
Akata \etal \cite{Akata2015} & - & 61.9 & \textbf{\em 40.3} & - & - \\
Lampert \etal \cite{10.1109/TPAMI.2013.140} & 38.16 & 57.23 & - & 72.00 & - \\
Romera-Paredes and Torr \cite{Romera-Paredes2015} & 24.22$\pm$2.89 & 75.32$\pm$2.28 & - & 82.10$\pm$0.32 & - \\
SSE-INT \cite{Zhang2015} & 44.15$\pm$0.34 & 71.52$\pm$0.79 & 30.19$\pm$0.59 & 82.17$\pm$0.76 & 57.01 \\
SSE-ReLU \cite{Zhang2015} & \textbf{\em 46.23$\pm$0.53} & \textbf{\em 76.33$\pm$0.83} & 30.41$\pm$0.20 & \textbf{\em 82.50$\pm$1.32} & \textbf{\em 58.87} \\
\hline\hline
(\rmnum{1}) init. $\forall \mathbf{z}_i^{(s)}, \forall \mathbf{z}_j^{(t)}$ + init. $\forall \mathbf{z}_{i'}^{(s)}, \forall \mathbf{z}_{j'}^{(t)}$ + Eq. \ref{eqn:decision1} & 38.10$\pm$2.64 & 76.96$\pm$1.40 & 39.03$\pm$0.87 & 81.17$\pm$2.02 & 58.81\\
(\rmnum{2}) init. $\forall \mathbf{z}_i^{(s)}, \forall \mathbf{z}_j^{(t)}$ + init. $\forall \mathbf{z}_{i'}^{(s)}, \forall \mathbf{z}_{j'}^{(t)}$ + Eq. \ref{eqn:decision2} & 38.20$\pm$2.75 & 80.11$\pm$1.13 & 41.07$\pm$0.81 & 81.33$\pm$1.76 & 60.20\\
(\rmnum{3}) init. $\forall \mathbf{z}_i^{(s)}, \forall \mathbf{z}_j^{(t)}$ + Alg. \ref{alg:test} + Eq. \ref{eqn:decision1} & 47.29$\pm$1.45 & 74.92$\pm$2.51 & 38.94$\pm$0.81 & 80.67$\pm$2.57 & 60.46\\
(\rmnum{4}) init. $\forall \mathbf{z}_i^{(s)}, \forall \mathbf{z}_j^{(t)}$ + Alg. \ref{alg:test} + Eq. \ref{eqn:decision2} & 47.79$\pm$1.83 & 77.37$\pm$0.39 & 40.91$\pm$0.86 & 80.83$\pm$2.25 & 61.73\\
(\rmnum{5}) Alg. \ref{alg:train} + init. $\forall \mathbf{z}_{i'}^{(s)}, \forall \mathbf{z}_{j'}^{(t)}$ + Eq. \ref{eqn:decision1} & 39.13$\pm$2.35 & 77.58$\pm$0.81 & 39.92$\pm$0.20 & 83.00$\pm$1.80 & 59.91\\
(\rmnum{6}) Alg. \ref{alg:train} + init. $\forall \mathbf{z}_{i'}^{(s)}, \forall \mathbf{z}_{j'}^{(t)}$ + Eq. \ref{eqn:decision2} & 38.94$\pm$2.27 & \textbf{\em 80.46$\pm$0.53} & \textbf{\em 42.11$\pm$0.55} & 82.83$\pm$1.61 & 61.09\\
(\rmnum{7}) Alg. \ref{alg:train} + Alg. \ref{alg:test} + Eq. \ref{eqn:decision1} & 50.21$\pm$2.90 & 76.43$\pm$0.75 & 39.72$\pm$0.19 & 83.67$\pm$0.29 & 62.51\\
(\rmnum{8}) Alg. \ref{alg:train} + Alg. \ref{alg:test} + Eq. \ref{eqn:decision2} & \textbf{\em 50.35$\pm$2.97} & 79.12$\pm$0.53 & 41.78$\pm$0.52 & \textbf{\em 83.83$\pm$0.29} & \textbf{\em 63.77}\\
\hline
\end{tabular}
\end{minipage}
\end{table*}

\section{Experiments}
We test our method on four benchmark image datasets for zero-shot recognition and retrieval, \ie aPascal \& aYahoo (aP\&Y) \cite{farhadi2009attribute}, Animals with Attributes (AwA) \cite{citeulike:7491128}, Caltech-UCSD Birds-200-2011 (CUB-200-2011) \cite{WahCUB_200_2011}, and SUN Attribute \cite{patterson2014sun}. Table~\ref{tab:dataset} summarizes the statistics in each dataset. In our experiments we utilized the same experimental settings as \cite{Zhang2015}. For comparison purpose we report our results averaged over 3 trials\footnote{Our code and CNN features can be downloaded at \url{https://zimingzhang.wordpress.com/}.}.

\subsection{Implementation}\label{ssec:implementation}
\noindent
{\bf (\rmnum{1}) Cross validation:} Similar to \cite{Zhang2015}, we utilize cross validation to tune the parameters. Precisely, we randomly select two seen classes from training data for validation purpose, train our method on the rest of the seen classes, and record the performance using different parameter combinations. We choose the parameters with the best average performance on the held-out seen class data.

\noindent
{\bf (\rmnum{2}) Dictionary initialization:} For source domain, we initialize the dictionary $\mathbf{B}$ to be the collection of all the seen class attribute vectors 
on aP\&Y, AwA, and CUB-200-2011, because of the paucity of the number of vectors. On SUN, however, for computational reasons, we initialize $\mathbf{B}$ using KMeans with 200 clusters on the attribute vectors. 

\begin{figure}[t]
\begin{minipage}[b]{0.495\columnwidth}
 \begin{center}
 \centerline{\includegraphics[width=.9\columnwidth]{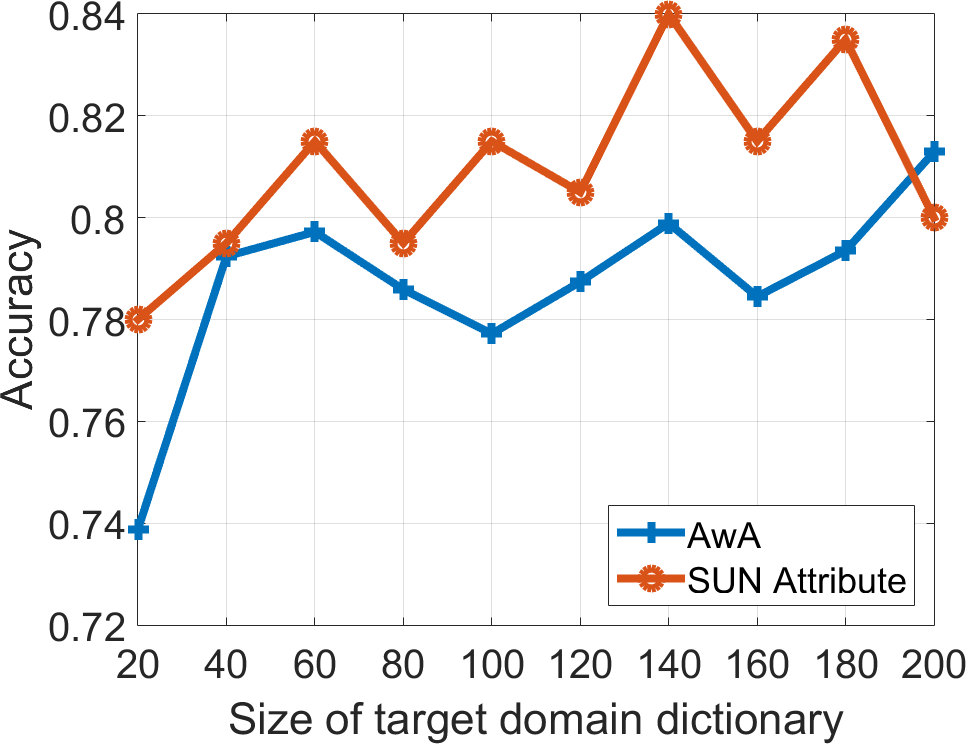}}
 \centerline{\footnotesize{(a)}}
 \end{center}
\end{minipage}
\begin{minipage}[b]{0.495\columnwidth}
 \begin{center}
 \centerline{\includegraphics[width=.9\columnwidth]{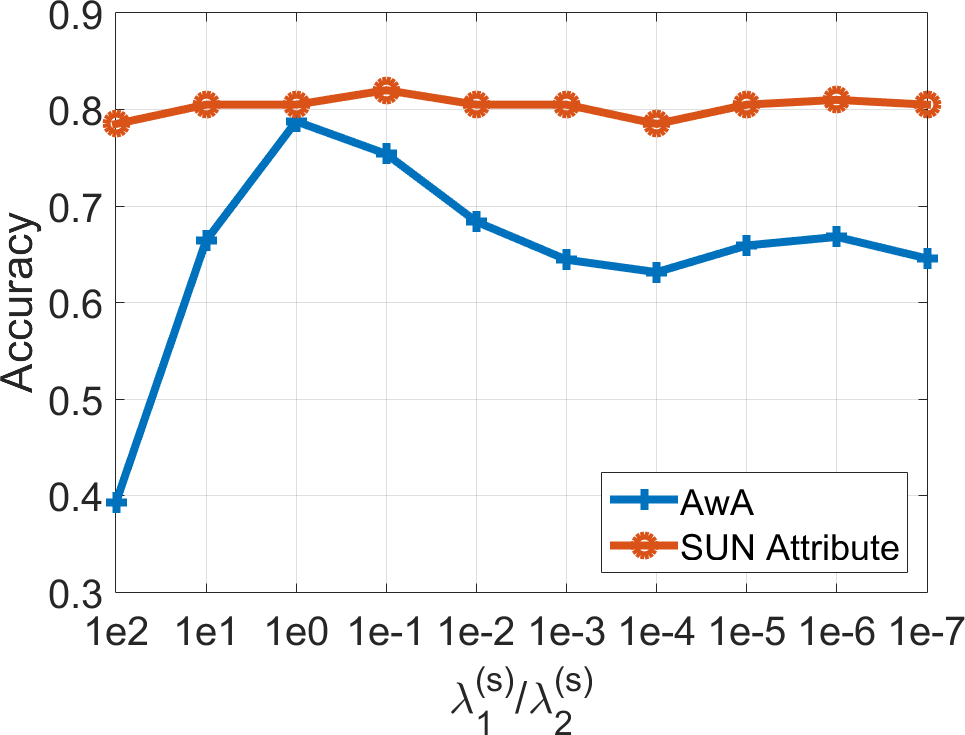}}
 \centerline{\footnotesize{(b)}}
 \end{center}
\end{minipage}
\vspace{-7mm}
\caption{\footnotesize{Effect of {\bf (a)} the size of target domain dictionary, and {\bf (b)} source domain parameter ratio $\lambda_1^{(s)}/\lambda_2^{(s)}$ on accuracy.}}\label{fig:implementation}
\vspace{-3mm}
\end{figure}

For target domain, we utilize the top eigenvectors of all training data samples to initialize the dictionary $\mathbf{D}$. In Fig. \ref{fig:implementation}(a), we show the effect of varying the size of $\mathbf{D}$ on our accuracy on AwA and SUN Attribute datasets. As we see, within small ranges of dictionary size, our performance changes marginally. We set the initial sizes to be 40, 200, 300, and 200, for the four datasets respectively, and then tune them using cross validation.

\noindent
{\bf (\rmnum{3}) Regularization parameters in Eq. \ref{eqn:pb}, \ref{eqn:pd}, and \ref{eqn:pw}:} We do a grid search to tune these parameters. In order to show how well our method adapts to different parameters, we display salient results in Fig. \ref{fig:implementation}(b), for varying source domain parameter ratios ($\lambda_1^{(s)}/\lambda_2^{(s)}$) on AwA and SUN datasets. 

\begin{figure*}[t]
\begin{minipage}[b]{0.495\columnwidth}
 \begin{center}
 \centerline{\includegraphics[width=.9\columnwidth]{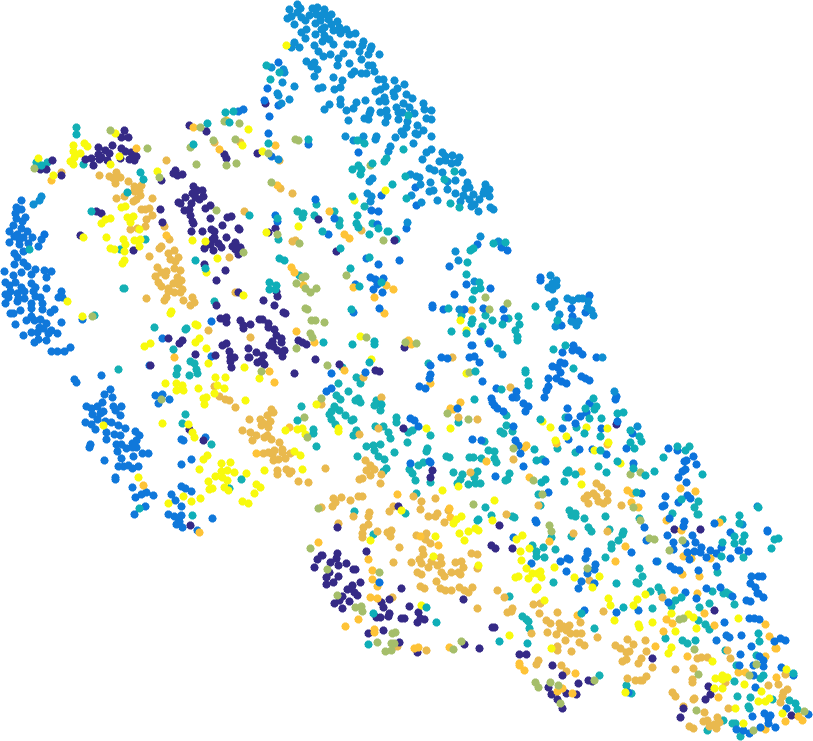}}
 \centerline{\footnotesize{(a) SSE: decaf}}
 \end{center}
\end{minipage}
\begin{minipage}[b]{0.495\columnwidth}
 \begin{center}
 \centerline{\includegraphics[width=.9\columnwidth]{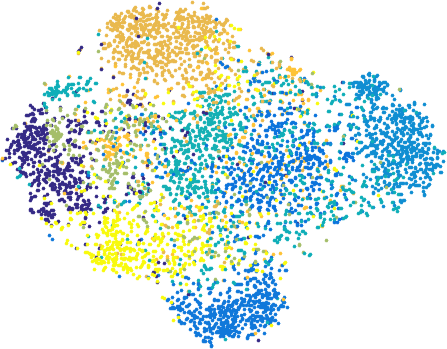}}
 \centerline{\footnotesize{(b) Ours: decaf}}
 \end{center}
\end{minipage}
\begin{minipage}[b]{0.495\columnwidth}
 \begin{center}
 \centerline{\includegraphics[width=.9\columnwidth]{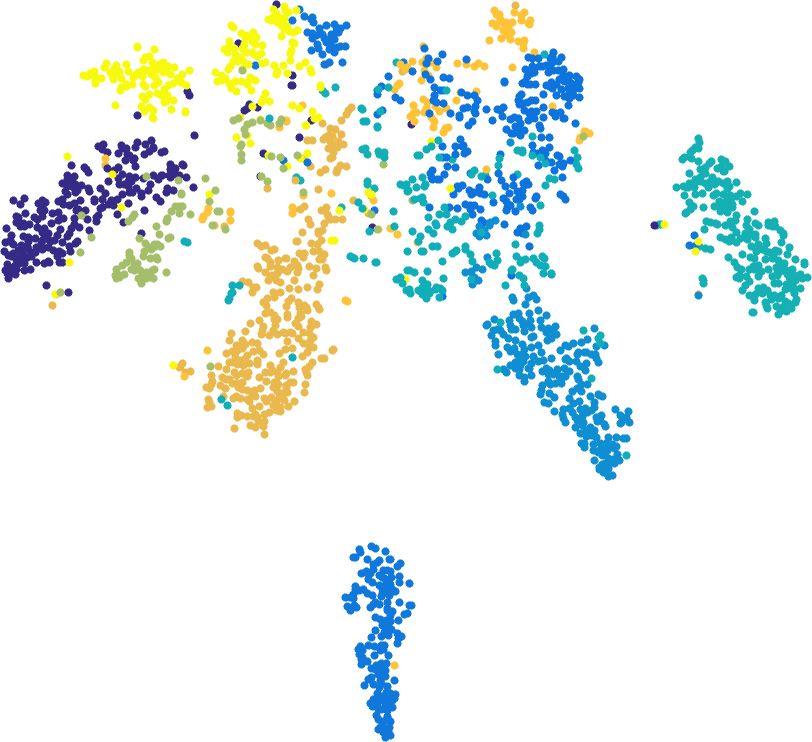}}
 \centerline{\footnotesize{(c) SSE: verydeep-19}}
 \end{center}
\end{minipage}
\begin{minipage}[b]{0.495\columnwidth}
 \begin{center}
 \centerline{\includegraphics[width=.9\columnwidth]{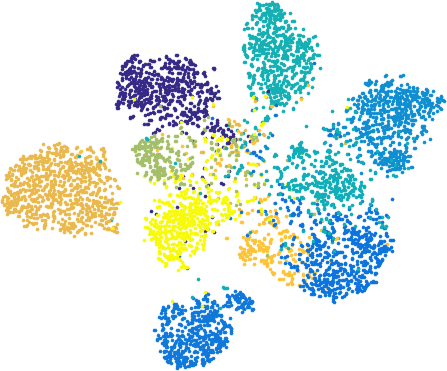}}
 \centerline{\footnotesize{(d) Ours: verydeep-19}}
 \end{center}
\end{minipage}
\vspace{-3mm}
\caption{\footnotesize{t-SNE visualization comparison between {\bf (a, c)} SSE \cite{Zhang2015} and {\bf (b, d)} our method using decaf and verydeep-19 features on AwA testing data from unseen classes, respectively. Clearly our method can better separate features from different classes.}}\label{fig:awa-feat}
\end{figure*}

\begin{figure*}[t]
\begin{minipage}[b]{0.195\linewidth}
 \begin{center}
 \centerline{\includegraphics[width=.75\columnwidth]{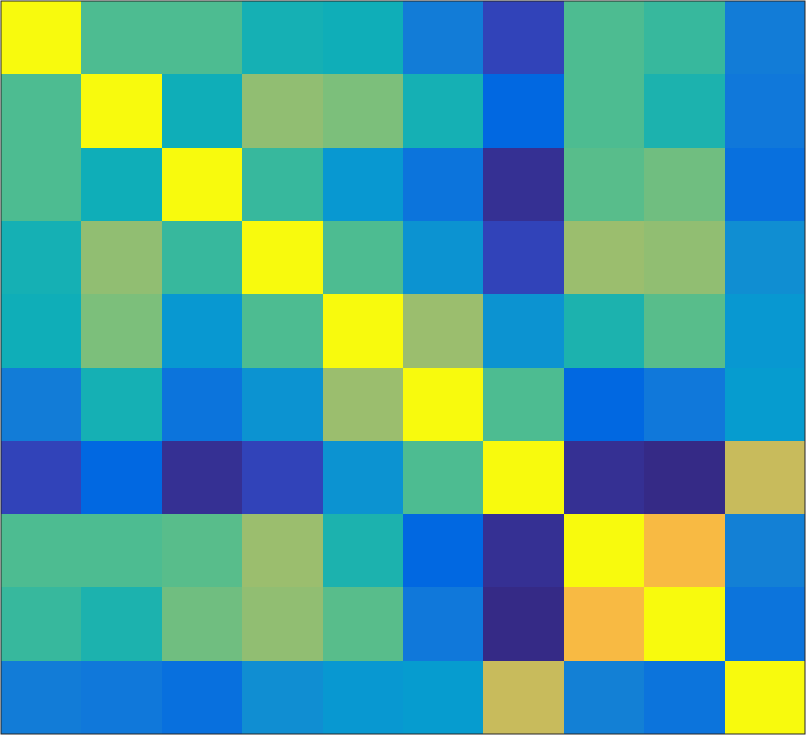}}
 \centerline{\footnotesize{(a) Attributes}}
 \end{center}
\end{minipage}
\begin{minipage}[b]{0.195\linewidth}
 \begin{center}
 \centerline{\includegraphics[width=.75\columnwidth]{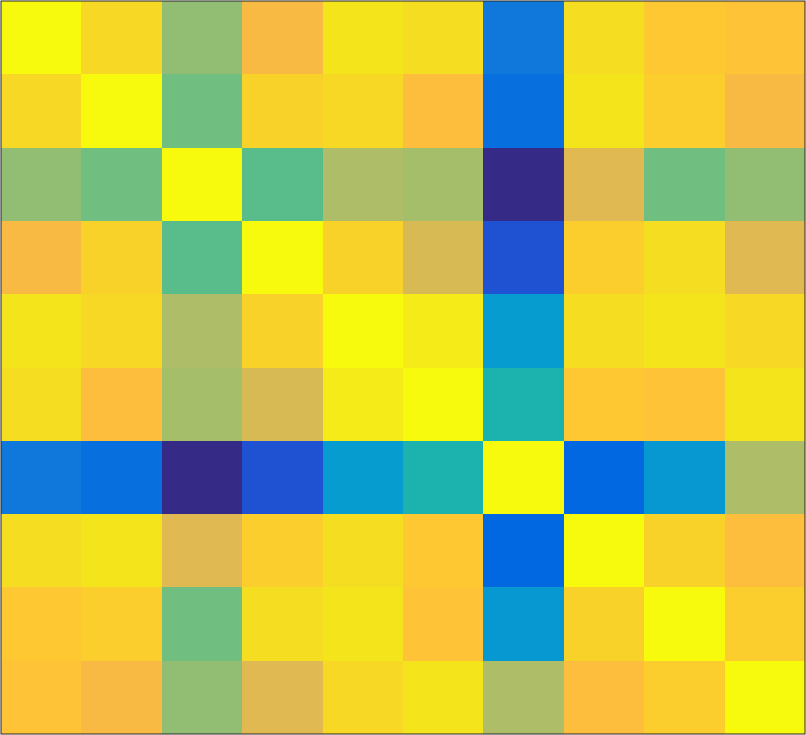}}
 \centerline{\footnotesize{(b) SSE: decaf}}
 \end{center}
\end{minipage}
\begin{minipage}[b]{0.195\linewidth}
 \begin{center}
 \centerline{\includegraphics[width=.75\columnwidth]{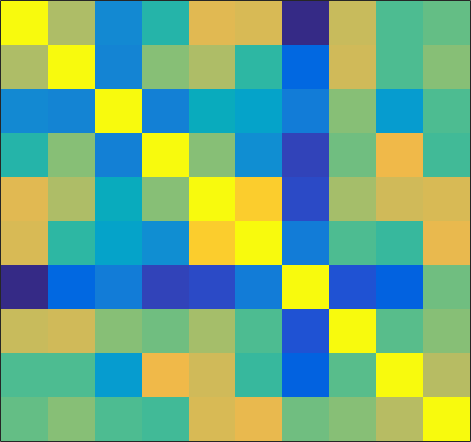}}
 \centerline{\footnotesize{(c) Ours: decaf}}
 \end{center}
\end{minipage}
\begin{minipage}[b]{0.195\linewidth}
 \begin{center}
 \centerline{\includegraphics[width=.75\columnwidth]{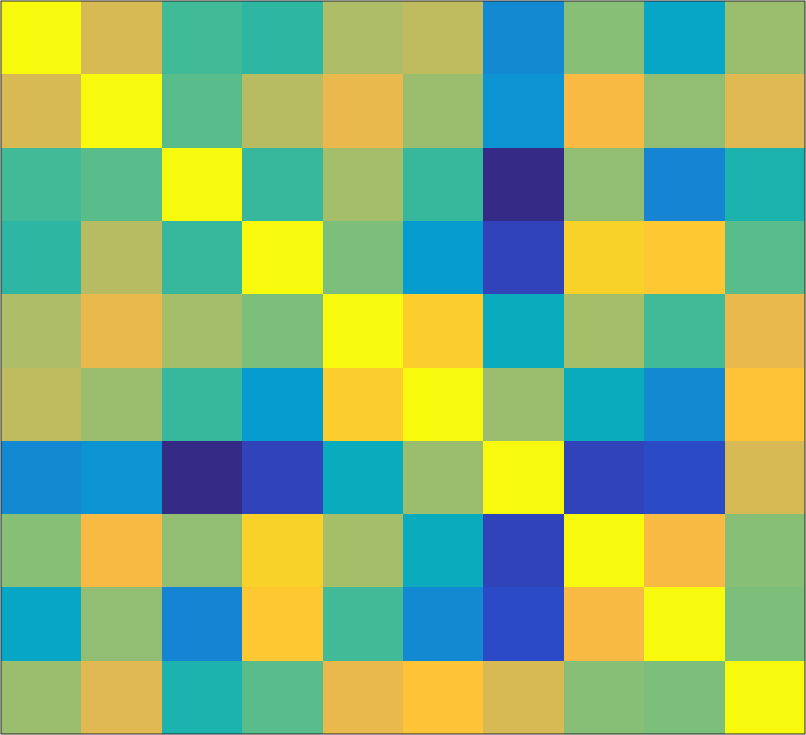}}
 \centerline{\footnotesize{(d) SSE: verydeep-19}}
 \end{center}
\end{minipage}
\begin{minipage}[b]{0.195\linewidth}
 \begin{center}
 \centerline{\includegraphics[width=.75\columnwidth]{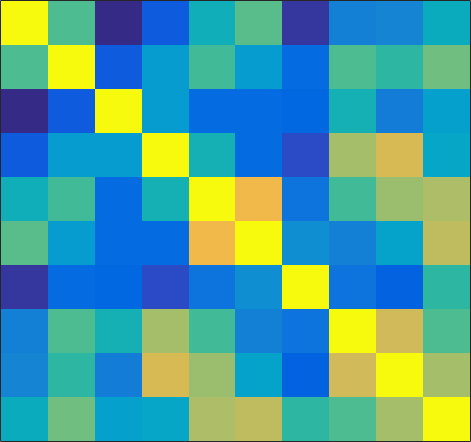}}
 \centerline{\footnotesize{(e) Ours: verydeep-19}}
 \end{center}
\end{minipage}
\vspace{-3mm}
\caption{\footnotesize{Comparison of cosine similarity matrices created using different features on AwA testing data using {\bf (a)} source domain attribute vectors, {\bf (b, d)} SSE \cite{Zhang2015} with decaf and verydeep-19, and {\bf (c, e)} our method with decaf and verydeep-19, respectively. Brighter colors depict larger values.}}\label{fig:awa-mat}
\vspace{-3mm}
\end{figure*}

\subsection{Benchmark Comparison}
On the four datasets, we perform two different tasks: (1) {\em zero-shot recognition} and (2) {\em zero-shot retrieval}. While both tasks are related, they measure different aspects of the system. Task 1 is fundamentally about classification of each target data instance. Task 2 measures which target domain samples are matched to a given source domain vector, and we adapt our recognition system for the purpose of retrieval. Specifically, given a source domain unseen class attribute vector we compute the similarities for all the unseen target domain data and sort the similarity scores. We can then compute precision, recall, average precision (AP) etc. to measure retrieval accuracy.
%

\subsubsection{Zero-Shot Recognition}
Recognition accuracy for each method is presented in Table~\ref{tab:benchmark}. We also perform an ablative study in order to understand the contribution of different parts of our system. We experiment with the three parts of our system: (1) dictionary learning; (2) test-time latent variable estimation; (3) incorporating source domain data fit term in prediction.

%
Note that the source and target domain dictionaries $\mathbf{B}$ and $\mathbf{D}$ are initialized in the beginning of the dictionary learning process (see Sec 3.1 (ii)). Consequently, we can bypass dictionary learning (deleting $repeat$ loop in Alg \ref{alg:train}) and understand its impact. Next we can ignore the similarity function term for estimating the latent embeddings for unseen data during test-time.
Finally, we can choose one of the two prediction rules (Eq.~\ref{eqn:decision1} or Eq.~\ref{eqn:decision2}) to determine the utility of using source domain data fit term for prediction. 
We denote by 
``init. $\forall \mathbf{z}_{i}^{(s)}, \forall \mathbf{z}_{j}^{(t)}$'' when dictionary learning is bypassed; We denote by ``init. $\forall \mathbf{z}_{i'}^{(s)}, \forall \mathbf{z}_{j'}^{(t)}$'' when similarity term is ignored during test-time. 
We list all the 8 choice combinations for our system in Table \ref{tab:benchmark} (\rmnum{1}) to (\rmnum{8}). 

The overall best result is obtained for the most complex system using all parts of our system.
For instance, as seen from (\rmnum{1}) and (\rmnum{7}) we can see 3.70\% gain in average recognition accuracy. Our algorithm ``(\rmnum{8}) Alg.~\ref{alg:train} + Alg.~\ref{alg:test} + Eq.~\ref{eqn:decision2}'' achieves the best result among all the competitors, significantly outperforming the state-of-the-art by 4.90\%. In the rest of the paper, we refer to (\rmnum{8}) as our method by default.
Table \ref{tab:benchmark} also demonstrates that on average, (a) the decision function in Eq.~\ref{eqn:decision2} performs better than that in Eq.~\ref{eqn:decision1}, and (b) test-time learning of unseen class latent embeddings using Alg.~\ref{alg:test} is more important than dictionary learning. For instance, by comparing (\rmnum{1}) with (\rmnum{2}), using Eq.~\ref{eqn:decision2} the performance gains are 1.39\% improvement over Eq.~\ref{eqn:decision1}. We see modest gains (0.55\%) from (\rmnum{3}) to (\rmnum{5}). 
Still our ablative study demonstrates that on individual datasets there is no single system that dominates other system-level combinations. Indeed, for aP\&Y (\rmnum{6}) is worse than (\rmnum{5}).

We visually depict (see Fig. \ref{fig:awa-feat}) the learned test-time unseen class embeddings, using t-SNE \cite{van2008visualizing} on AwA to facilitate better understanding of our results with respect to the state-of-art \cite{Zhang2015}. 
Our method appears to learn more separable embeddings regardless of the target domain features (decaf \cite{donahue2014decaf} or verydeep-19). Indeed, as seen in Fig. \ref{fig:awa-feat} (b,d) the embeddings appear to be more cluttered than those in (a,c).

Next, in Fig. \ref{fig:awa-mat} we plot the cosine similarity matrices for the learned embeddings as in \cite{Zhang2015} on the AwA dataset. Note that \cite{Zhang2015} employs so called semantic similarity embedding (SSE). 
The figures demonstrate that our method can generate a cosine similarity matrix which is much more similar to the source domain attribute cosine similarity (a).
Fig. \ref{fig:awa-feat} and Fig. \ref{fig:awa-mat} together demonstrate that our method is capable of aligning the source and target domain data better than the state-of-the-art method \cite{Zhang2015}. In addition it is capable of learning qualitatively better (clustered) embedding representations for different classes, leading to improvements in recognition accuracy on the four benchmark datasets.

\subsubsection{Zero-Shot Retrieval}

\begin{figure*}[t]
\begin{minipage}[b]{0.325\textwidth}
 \begin{center}
 \centerline{\includegraphics[width=.93\columnwidth]{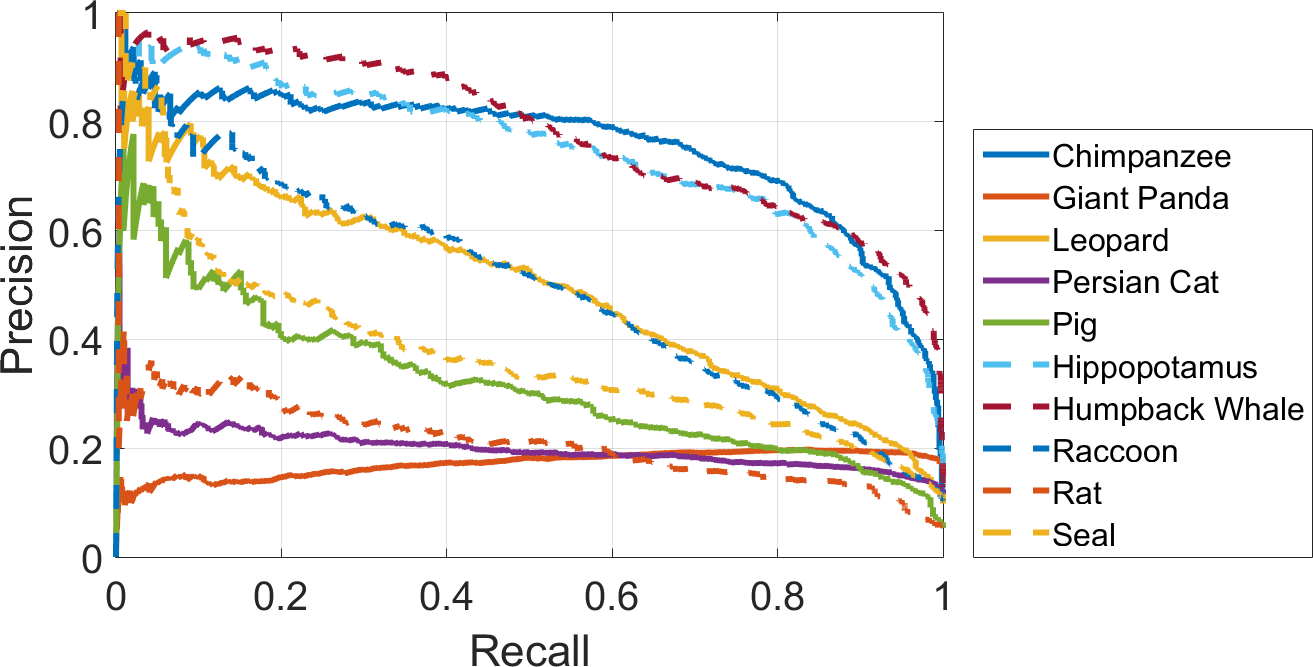}}
 \centerline{\footnotesize{(a) SSE-INT}}
 \end{center}
\end{minipage}
\begin{minipage}[b]{0.325\textwidth}
 \begin{center}
 \centerline{\includegraphics[width=.93\columnwidth]{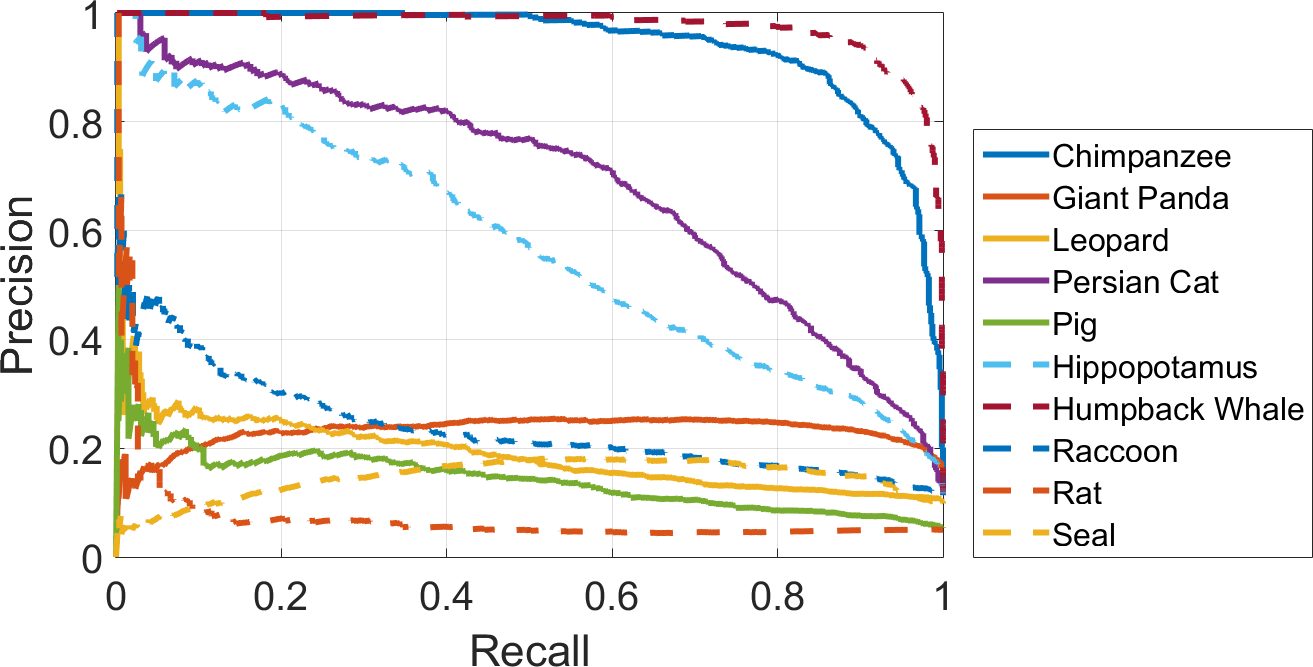}}
 \centerline{\footnotesize{(b) SSE-ReLU}}
 \end{center}
\end{minipage}
\begin{minipage}[b]{0.325\textwidth}
 \begin{center}
 \centerline{\includegraphics[width=.93\columnwidth]{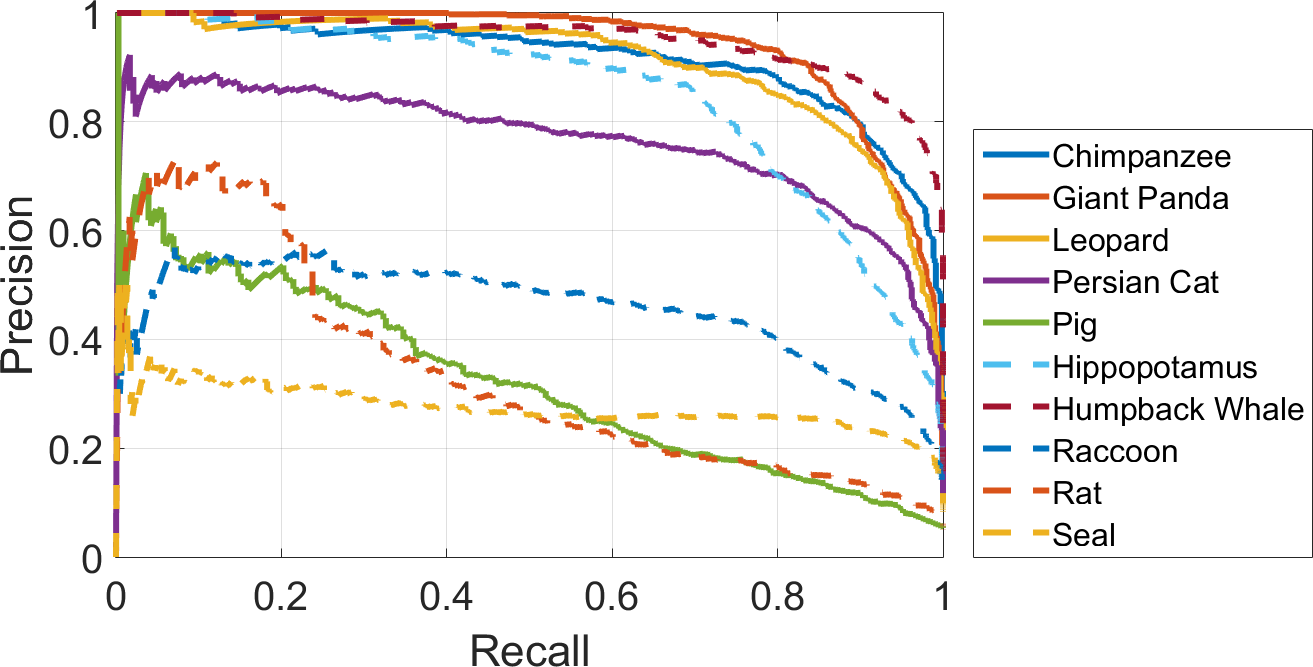}}
 \centerline{\footnotesize{(c) Ours}}
 \end{center}
\end{minipage}
\vspace{-4mm}
\caption{\footnotesize{Illustration of precision-recall curve comparison on AwA.}}
\label{fig:retrieval}
\vspace{-2mm}
\end{figure*}

\begin{figure}[t]
\centerline{\includegraphics[width=.9\linewidth]{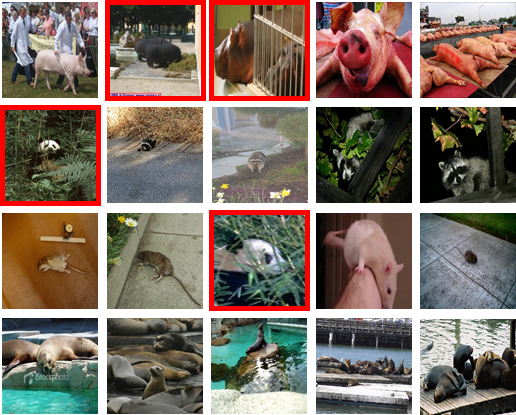}}
\vspace{2mm}
\caption{\footnotesize{Top-5 zero-shot retrieval results using our method for class {\bf (from top to down)} ``Pig'', ``Raccoon'', ``Rat'', and ``Seal'', respectively. Images with red rectangles are false-positive returns.}}\label{fig:awa_samples}
\vspace{-5mm}
\end{figure}

%

\begin{table*}[t]\footnotesize
\begin{minipage}[b]{0.4\textwidth}
 \begin{center}
\setlength\tabcolsep{3pt}
\caption{\footnotesize{Retrieval performance comparison (\%) using mAP.}}\label{tab:retrieval}\vspace{1mm}
\begin{tabular}{|l|llll|l|}
\hline
Method & aP\&Y & AwA & CUB & SUN & Ave. \\
\hline\hline  
SSE-INT \cite{Zhang2015} & 15.43 & 46.25 & 4.69 & 58.94 & 31.33 \\
SSE-ReLU \cite{Zhang2015} & 14.09 & 42.60 & 3.70 & 44.55 & 26.24 \\
\hline\hline  
Ours & \textbf{\em 38.30} & \textbf{\em 67.66} & \textbf{\em 29.15} & \textbf{\em 80.01} & \textbf{\em 53.78} \\
\hline
\end{tabular}
 \end{center}
\end{minipage}
\begin{minipage}[b]{0.6\textwidth}
 \begin{center}
\setlength\tabcolsep{3pt}
\caption{\footnotesize{Retrieval performance comparison (\%) using AP on AwA.}}\label{tab:retrieval_awa}\vspace{1mm}
\begin{tabular}{|llllllllll|l|}
\hline
Chim. & Panda & Leop. & Cat & Pig & Hipp. & Whale & Racc. & Rat & Seal & mAP \\
\hline\hline  
76.05 & 19.67 & 50.12 & 20.33 & 32.83 & 74.88 & 78.31 & \textbf{\em 50.52} & 21.85 & \textbf{\em 37.96} & 46.25 \\
\textbf{\em 94.20} & 24.81 & 19.24 & 69.08 & 14.73 & 57.51 & \textbf{\em 97.56} & 24.11 & 7.59 & 17.20 & 42.60 \\
\hline\hline  
91.75 & \textbf{\em 94.06} & \textbf{\em 91.09} & \textbf{\em 76.95} & \textbf{\em 33.00} & \textbf{\em 84.85} & 95.13 & 47.05 & \textbf{\em 34.58} & 28.18 & \textbf{\em 67.66} \\
\hline
\end{tabular}
 \end{center}
\end{minipage}
\vspace{-4mm}
\end{table*}

We list comparative results for the mean average precision (mAP) for the four datasets in Table \ref{tab:retrieval}. Since retrieval is closely related to recognition and, SSE \cite{Zhang2015} is the state-of-art, we focus on comparisons with it. As we can see our method significantly and consistently outperforms SSE by 22.45\% on average. Our superior performance in retrieval is due to the better domain alignment and more clustered embedding representations. This leads to better matching of target domain data to source domain vectors. 
Our retrieval results are based on adapting the recognition models for the retrieval task. It is possible that incorporating pairwise ranking constraints into the training (\eg into Eq. \ref{eqn:pw} for our method) may improve performance, but it is outside the scope of this paper. 

We again attempt to further analyze our method on the AwA dataset. We list class-wise AP as well as mAP comparison in Table \ref{tab:retrieval_awa}, and illustrate the precision-recall curves for different methods in Fig. \ref{fig:retrieval}. Our method achieves over 70\% AP for 6 out of 10 classes, and performs the best in 6 out of 10 classes. Fig. \ref{fig:retrieval} depicts illustrative examples for different categories. Nevertheless, we note that for some classes our method is unable to achieve satisfactory performance (although other methods also suffer from performance degradation). For instance, we only get 28.18\% AP for class ``seal''. Note that in Fig. \ref{fig:awa-mat}(e), we can see that the last row (or column), which corresponds to ``seal'', shows some relatively high values in off-diagonal elements. This is because the problem of differentiating data within this class from data from other classes is difficult. Similar situations can be observed in SSE as well.

We also visualize our retrieval results in Fig. \ref{fig:awa_samples} with the top-5 returns for ``difficult'' cases (classes with AP less than 50\%) in Table \ref{tab:retrieval_awa}. Interestingly for the most difficult class ``seal'', all five images are correct. This is probably because the global patterns such as texture in the images are similar, leading to highly similar yet discriminative CNN features.

\section{Conclusion}
In this paper we propose a novel general probabilistic method for ZSL by learning joint latent similarity embeddings for both source and target domains. Based on the equivalence of ZSR and binary prediction, and the conditional independence between observed data and predicted class, we propose factorizing the likelihood of binary prediction using our probabilistic model to jointly learn the latent spaces for each domain. In this way, we generate a joint latent space for measuring the latent similarity between source and target data. Our similarity function is invariant across different classes, and hence intuitively it fits well to ZSR with good generalization to unseen classes. We further propose a new supervised dictionary learning based ZSR algorithm as parametrization of our probabilistic model. We conduct comprehensive experiments on four benchmark datasets for ZSL with two different tasks, \ie zero-shot recognition and retrieval. We evaluate the importance of each key component in our algorithm, and show significant improvement over the state-of-the-art. Possible applications are person re-identification \cite{zhang_eccv14,zhang_iccv15_reid,zhang2014person} and zero-shot activity retrieval \cite{Castanon_mm15}.

\section*{Acknowledgement}\small
We thank the anonymous reviewers for their very useful comments. This material is based upon work supported in part by the U.S. Department of Homeland Security, Science and Technology Directorate, Office of University Programs, under Grant Award 2013-ST-061-ED0001, by ONR Grant 50202168 and US AF contract FA8650-14-C-1728. The views and conclusions contained in this document are those of the authors and should not be interpreted as necessarily representing the social policies, either expressed or implied, of the U.S. DHS, ONR or AF.

\newpage

{\footnotesize
\bibliographystyle{ieee}
\bibliography{egbib}
}

\end{document}